\documentclass{article}

\usepackage[authoryear]{natbib}  % or [authoryear] for author-year citations

\usepackage[final]{neurips_2025}

\usepackage[utf8]{inputenc} % allow utf-8 input
\usepackage[T1]{fontenc}    % use 8-bit T1 fonts
\usepackage[colorlinks=true,      % false = boxed links
            linkcolor=red,       % color of internal links (e.g. section refs)
            citecolor=blue,      % color of citations
            filecolor=magenta,   % color of file links
            urlcolor=cyan        % color of external links
            ]{hyperref}
\usepackage{url}            % simple URL typesetting
\usepackage{booktabs}       % professional-quality tables
\usepackage{amsfonts}       % blackboard math symbols
\usepackage{amsmath}       % 
\usepackage{nicefrac}       % compact symbols for 1/2, etc.
\usepackage{microtype}      % microtypography
\usepackage{xcolor}         % colors
\usepackage{graphicx}
\usepackage[capitalise]{cleveref}       % cref
\crefname{section}{\S\!}{\S\!}
\usepackage{placeins}
\usepackage{subcaption}
\usepackage[normalem]{ulem}
\usepackage{wrapfig}
\usepackage{float}
\usepackage{enumitem}

\usepackage{float}

\newif\ifdraft
\drafttrue
\ifdraft
\newcommand{\argocomment}[1]{{\color{brown}[\textbf{Argo:} #1]}}
\newcommand{\gpc}[1]{{\color{blue}[\textbf{Gaurav:} #1]}}

\newcommand{\kac}[1]{{\color{purple}[\textbf{Kfir:} #1]}}
\newcommand{\opc}[1]{{\color{red}[\textbf{Or:} #1]}}
\newcommand{\dcc}[1]{{\color{orange}[\textbf{Danny:} #1]}}
\newcommand{\jacksoncomment}[1]{{\color{teal}[\textbf{Jackson:} #1]}}
\newcommand{\jacksonc}[1]{{\color{teal}[\textbf{Jackson:} #1]}}
\newcommand{\dosc}[1]{{\color{magenta}[\textbf{Daniil:} #1]}}
\newcommand{\jackson}[1]{{\color{teal}#1}}
\newcommand{\dos}[1]{{\color{magenta}#1}}
\else
\newcommand{\argocomment}[1]{}
\newcommand{\gpc}[1]{}
\newcommand{\kac}[1]{}
\newcommand{\opc}[1]{}
\newcommand{\dcc}[1]{}
\newcommand{\jacksoncomment}[1]{}
\newcommand{\jacksonc}[1]{}
\newcommand{\dosc}[1]{}

\newcommand{\jackson}[1]{}
\newcommand{\dos}[1]{}

\fi

\title{Preventing Shortcuts in Adapter Training via Providing the Shortcuts}

\newcommand{\webpage}{\url{https://snap-research.github.io/shortcut-rerouting/}~}

\author{%
Anujraaj Argo Goyal\quad 
Guocheng Gordon Qian \thanks{Corresponding authors: gqian@snapchat.com, jwang23@snapchat.com} \quad
Huseyin Coskun \quad
Aarush Gupta  \AND
Himmy Tam \quad
Daniil Ostashev \quad
Ju Hu \quad
Dhritiman Sagar  \AND
Sergey Tulyakov \quad
Kfir Aberman \quad
Kuan-Chieh Jackson Wang $^{*}$ \\
\\
Snap Inc., \texttt{\webpage}
}

\begin{document}

\maketitle

% \begin{figure}[h] 
% \centering
% \includegraphics[width=\textwidth]{figures/src/first-fig.002.png}
% \caption{
% We introduce \textit{Shortcut-Rerouted Adapter Training} (\textbf{SR-Training}) to address the artifacts introduced by distribution shift (e.g., degraded visual quality, distorted anatomy) and spurious correlation (e.g., same expression, head orientation, body pose as inputs) in personalization training. \argo{todo: cherry-pick examples and update caption, cloth leakage?\gordon{1. make it a single line as before. 2. the resolution is too low. 3. I like lelf top and bottom right, as they show better pose alignment and better background.}}
% \jacksonc{This teaser is too application focused.  Our paper should be about the abstract idea of shortcut rerouting.  I don't think we need this teaser for the paper.}

% }
% \label{fig:teaser}
% \end{figure}
\begin{abstract}
Adapter-based training has emerged as a key mechanism for extending the capabilities of powerful foundation image generators, enabling personalized and stylized text-to-image synthesis. These adapters are typically trained to capture a specific target attribute, such as subject identity, using single-image reconstruction objectives. However, because the input image inevitably contains a mixture of visual factors, adapters are prone to entangle the target attribute with incidental ones, such as pose, expression, and lighting. This spurious correlation problem limits generalization and obstructs the model's ability to adhere to the input text prompt.  In this work, we uncover a simple yet effective solution: provide the very shortcuts we wish to eliminate during adapter training. In \emph{Shortcut-Rerouted Adapter Training}, confounding factors are routed through auxiliary modules, such as ControlNet or LoRA, eliminating the incentive for the adapter to internalize them. The auxiliary modules are then removed during inference. When applied to tasks like facial and full-body identity injection, our approach improves generation quality, diversity, and prompt adherence. These results point to a general design principle in the era of large models: when seeking disentangled representations, the most effective path may be to establish shortcuts for what should \emph{not} be learned.
\end{abstract}

\section{Introduction}

In recent years, text-to-image (T2I) models have undergone remarkable progress, revolutionizing the way we generate and manipulate visual content from natural language prompts~\citep{ldm, ramesh2022hierarchical}. While the expressive power of these foundation models has unlocked myriad creative and practical applications, much of their flexibility is realized not through retraining the backbone itself, but through the introduction of lightweight \emph{adapters}. These adapters—ranging from low-rank adaptation modules (LoRAs)~\citep{hu2022lora} to encoders~\citep{ye2023ipadapter}—serve as modular steering mechanisms, enabling tailored functionality atop a frozen foundation model. LoRA-based adapters, for instance, have empowered stylized and user-preference-conditioned generation, while encoder-based adapters facilitate personalized synthesis and style injection with impressive specificity~\citep{ControlNet,Wang2023StyleAdapterAS,luo2024stylus}. In essence, adapter training has emerged as a key enabler of fine-grained control in the modern image generation landscape.

\begin{figure}[!t]
    \centering
    \includegraphics[width=\linewidth]{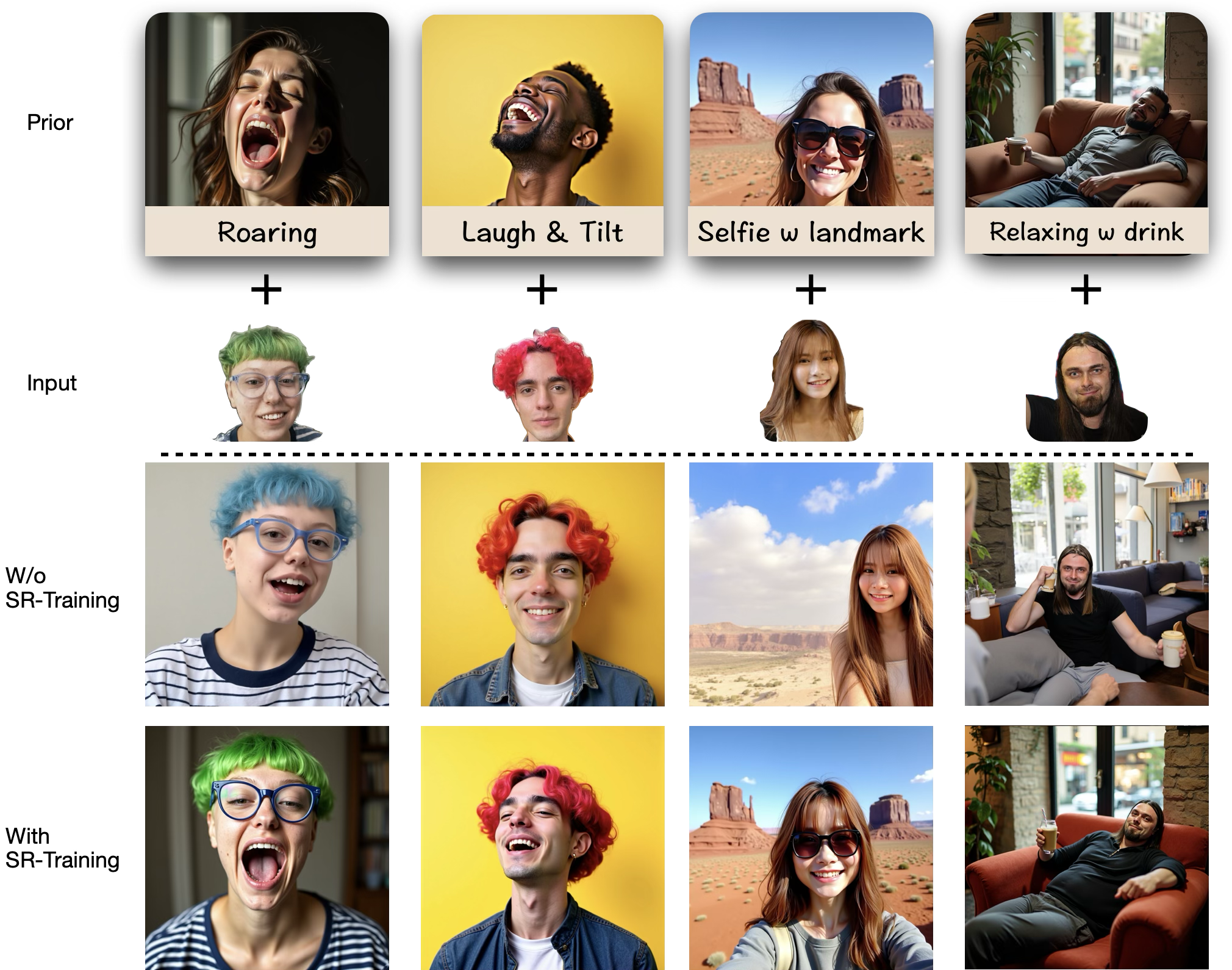}
    \caption{\textbf{Shortcut Rerouting re-enables text control of pose and expression after adapter training.} In the context of personalized generation, without shortcut rerouting, the adapter overfits to the reference image and reproduces its pose and expression, ignoring the prompt. With Shortcut-Rerouted Adapter Training, the adapter disentangles identity from other factors, allowing the model to respond faithfully to prompt-specified expressions and head poses. This restores compositionality, preserves the prior, and leads to more expressive and diverse generations.
    } 
    \label{fig:overview-of-results}
    \vspace{-1em}
\end{figure}

\begin{figure}[!t]
    \centering
    \includegraphics[width=\linewidth]{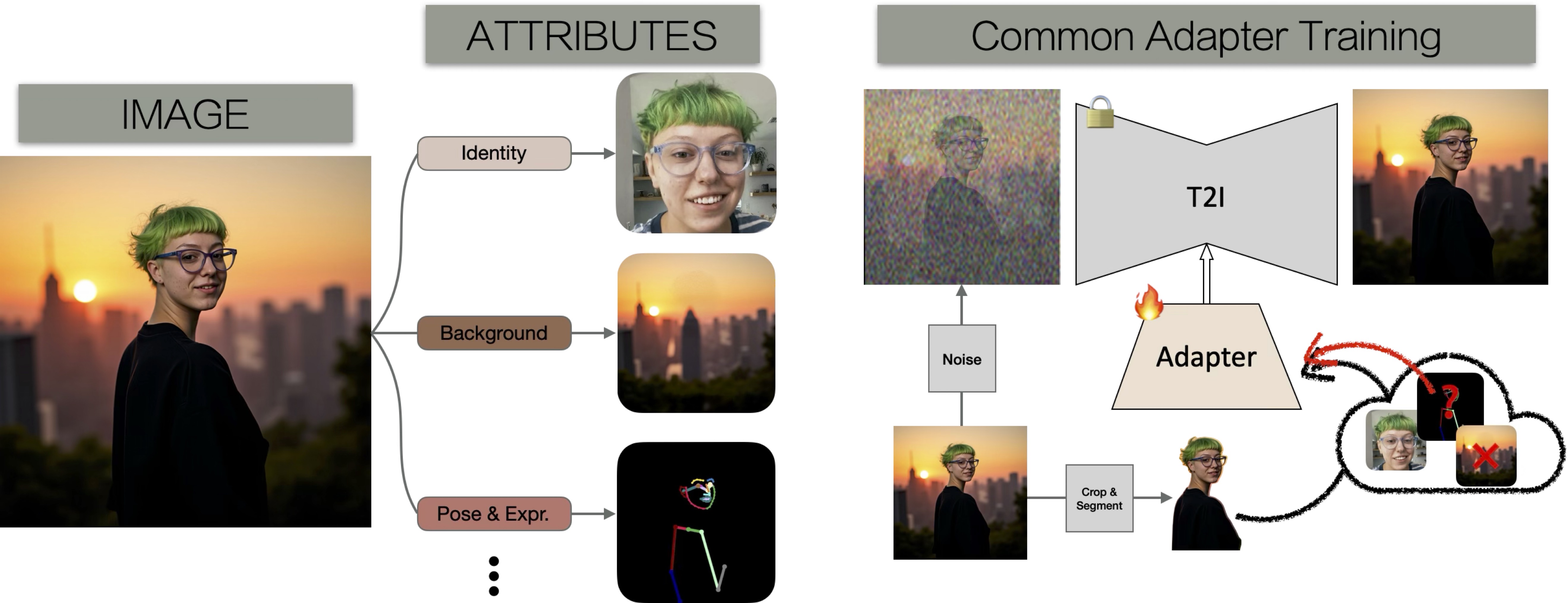}
    \caption{\textbf{Common adapter training is susceptible to learning undesired shortcutes.} The common \emph{single-image reconstruction} objective used in adapter training inadvertently encourages the adapter to pick up all the attributes in the adapter input (e.g. pose, expression, background, distribution) and leak them into the generation. While some confounding attributes like background can be factored out using masking, many other cannot. This makes learning a pure ``identity'' adapter challenging.
    } 
    \label{fig:intuition}
\end{figure}

Yet, adapters face a fundamental challenge inherent to their training paradigm. The predominant approach—\emph{single-image reconstruction}, thanks to its simplicity and scalability—asks the adapter to faithfully reproduce a target image from a conditioning signal. An image, as the adage goes, is worth a thousand words; more precisely, an image encodes an entire constellation of attributes—identity, style, geometry, camera parameters, lighting, and beyond. In most cases, however, we wish to learn to encode only some specific attributes, and a thousand words are simply too many. The reconstruction loss, being agnostic to this distinction, indiscriminately incentivizes the adapter to reproduce \emph{all} visual factors present in the image. As a consequence, the adapter entangles the target factor with myriad incidental ones (i.e. \emph{shortcuts}). See ~\cref{fig:intuition}. 
An identity adapter intended to inject only the subject/person's appearance undesirably also copies and pastes their expression, pose, and leaks lighting or background style. Nowhere is this conflation more problematic than in facial personalization, where isolating immutable appearance traits from mutable factors like head pose or expression proves difficult. Moreover, as another key compounding factor, the distribution of the finetuning dataset is often significantly different from that of the foundation model. Such  copy-and-paste adapter training often introduces artifacts, such as degraded background generation, distorted human anatomy, and reduced esthetic quality (see~\cref{fig:overview-of-results}). Ideally, an identity encoder must only inject identity.

% paragraph 3: summarize our key idea and contribution 
Our central idea is simple: to prevent the adapter from learning undesirable shortcuts, we explicitly \emph{provide} those shortcuts during training (\cref{fig:method}). Rather than hoping the adapter would disentangle complex factors on its own, we architect the learning process to \emph{route} incidental factors through auxiliary modules—thus relieving the adapter of the burden of accounting for them. This reshapes the optimization landscape: when components of the reconstruction target are already explained by dedicated controllers (e.g., respective modules handling distribution shifts, pose, or expression), the adapter has no incentive to duplicate that behavior. The result is a principled factorization of responsibility, wherein each module specializes in its designated role. 
Whereas a naive adapter tends to copy pose and expression directly from the input image—thereby reducing the fidelity of the model prior and leading to degraded background generation—an adapter trained with shortcut rerouting learns to inject only the target identity. This restores prompt-based control over pose and expression and improves prior preservation (\cref{fig:overview-of-results}). To that end, this approach not only improves compositionality but also enhances the overall realism and diversity of the generated images.

In summary, our \textbf{contributions} are as follows:
\begin{enumerate}
    \item We propose a simple yet effective training paradigm, \emph{Shortcut Rerouting}, for adapter training for large text-to-image (T2I) models. 
    \item We apply Shortcut Rerouting to the task of personalized image generation, addressing confounding factors such as distribution shift and spurious correlations. We demonstrate two practical instantiations of this idea using well-established tools—LoRA and ControlNet—to explicitly factor out these shortcuts.
    \item We empirically validate Shortcut-Rerouted adapters in two distinct settings—facial and full-body personalization—and show improved controllability (via text prompts) with respect to expression, head pose, and body pose, as well as stronger prior preservation. This leads to higher overall image quality, fidelity, and naturalness compared to several strong baselines.
\end{enumerate}

\section{Related Work}

\noindent\textbf{Adapters in T2I Generation \& Personalized Generation.}
The advent of text-to-image (T2I) diffusion models~\citep{ho2020denoising,ldm} has spurred a growing interest in modular methods for task-specific and personalized generation~\citep{ruiz2023dreambooth,gal2022image,voynov2023p+}. Central to this movement is the concept of \emph{adapters}—lightweight modules that steer the behavior of a frozen generative backbone. Among these, LoRA~\citep{hu2022lora} has emerged as a widely adopted technique, enabling fine-tuning via low-rank parameter updates.  For improved inference efficiency, encoder-based adapters, which inject conditioning signals through learned embeddings or attention modulation, have likewise been instrumental in enabling fine-grained control over appearance, style, and compositionality~\citep{ye2023ipadapter,wang2024moa,fastcomposer,qian2025omni,qian2025composeme}.
Yet, a well-known issue of adapter fine-tuning is that it entangles identity injection with other undesired attributes like style, lighting, pose, and expression.  In~\cref{sec:exp}, we apply Shortcut Rerouting to a simple adapter, IP-Adapter, and compare it to strong recent baselines including InfU~\citep{jiang2025infiniteyou}, PulID~\citep{PuLID}, and a community implementation of IP-Adapter. 

\noindent\textbf{Shortcuts \& Spurious Correlation.}
The phenomenon of \emph{shortcut learning}—where models exploit spurious or unintended correlations in the data to optimize the training objective—has been extensively studied in the broader machine learning literature~\citep{geirhos2020shortcut,luo2021rectifying}. In vision tasks, shortcuts often manifest when models rely on superficial cues, such as texture or background, instead of learning the intended high-level semantics~\citep{Geirhos19texure_bias}. Within the generative modeling community, recent works have noted analogous behaviors: generative models and their adapters may entangle target factors with irrelevant or transient features present in training data, leading to poor generalization and a lack of modularity.
Several approaches have been proposed to combat shortcut learning, including data augmentation~\citep{geirhos2020shortcut}, causal regularization~\citep{mart2019invariant_risk_mini}, and architectural interventions~\citep{Islam20PosInfo}. In the context of T2I generation, methods such as ControlNet~\citep{ControlNet} explicitly inject structural conditioning (e.g., pose, layout) to guide synthesis, offering a promising avenue for disentanglement. However, to our knowledge, no prior work has systematically leveraged such auxiliary modules during \emph{adapter training} to proactively absorb spurious factors. Our method bridges this gap by architecting a modular training process that reroutes undesired correlations through dedicated controllers, thereby preventing their entanglement in the adapter’s representation.

Lastly, the notion of employing stage-wise training—where an auxiliary “training LoRA” is used solely during training to mitigate distribution shift—has been explored in several prior works. For instance, \citet{jones2024customizing} disentangled style and content learning by first training a content LoRA and subsequently a style LoRA, using only the latter at inference to achieve clean style transfer. Similarly, \citet{guo2023animatediff} fine-tuned a LoRA on the final video dataset to better absorb the target distribution shift. \citet{ostrislora} proposed a LoRA variant capable of “un-distilling” Flux, allowing users to fine-tune the already step-distilled Flux[schnell] model. Building on these insights, our work generalizes this concept beyond LoRA-based adapters: we demonstrate that auxiliary modules such as ControlNet can likewise be trained to absorb spurious correlations—e.g., those related to pose or expression—thereby isolating shortcut factors from the main adapter’s representation.

\section{Shortcut-Rerouted Adapter Training}
\subsection{Mathematical Formalism}

We formalize adapter training within a probabilistic framework to clarify the core challenge of disentanglement. Let $X$ denote the observed image, which depends on two underlying factors: the \emph{target factor} $T$ (e.g., identity, style) and the \emph{confounding factors} $C$ (e.g., pose, distribution shift, expression). Formally, we assume:
\begin{equation}\label{eq:gen-model}
X \sim p(X \mid T, C),
\end{equation}

where $T$ is the factor we wish to faithfully capture via adapter training, and $C$ represents incidental attributes that are not of primary interest.

The adapter $\mathcal{A}$ is a function that takes $X$ as input and produces a representation $\mathcal{A}(X)$, which is used to steer the generative model $G$. Ideally, we seek:
\begin{equation}\label{eq:ideal-target}
G(\mathcal{A} (X )) \approx p(\cdot \mid T),
\end{equation}

meaning that the adapter should extract and inject only the information relevant to $T$, regardless of the confounds $C$.

However, in typical single-image reconstruction training (\cref{fig:intuition}), the objective is to minimize:
\begin{equation}\label{eq:recon-obj}
\mathbb{E}_{(X)}\big[\mathcal{L}\big(G(\mathcal{A}(X)), X\big)\big],
\end{equation}

which implicitly encourages $\mathcal{A}(X)$ to encode \emph{both} $T$ and $C$, since $X$ embodies all these factors. As a result, the adapter becomes entangled: instead of isolating the target factor $T$, it captures spurious correlations mediated by $C$.  These \emph{shortcuts}, by minimizing the objective through confounding factors, inadvertently become reinforced and impair generalization to test prompts.

\begin{figure}[!t]
    \centering
    \includegraphics[width=\linewidth]{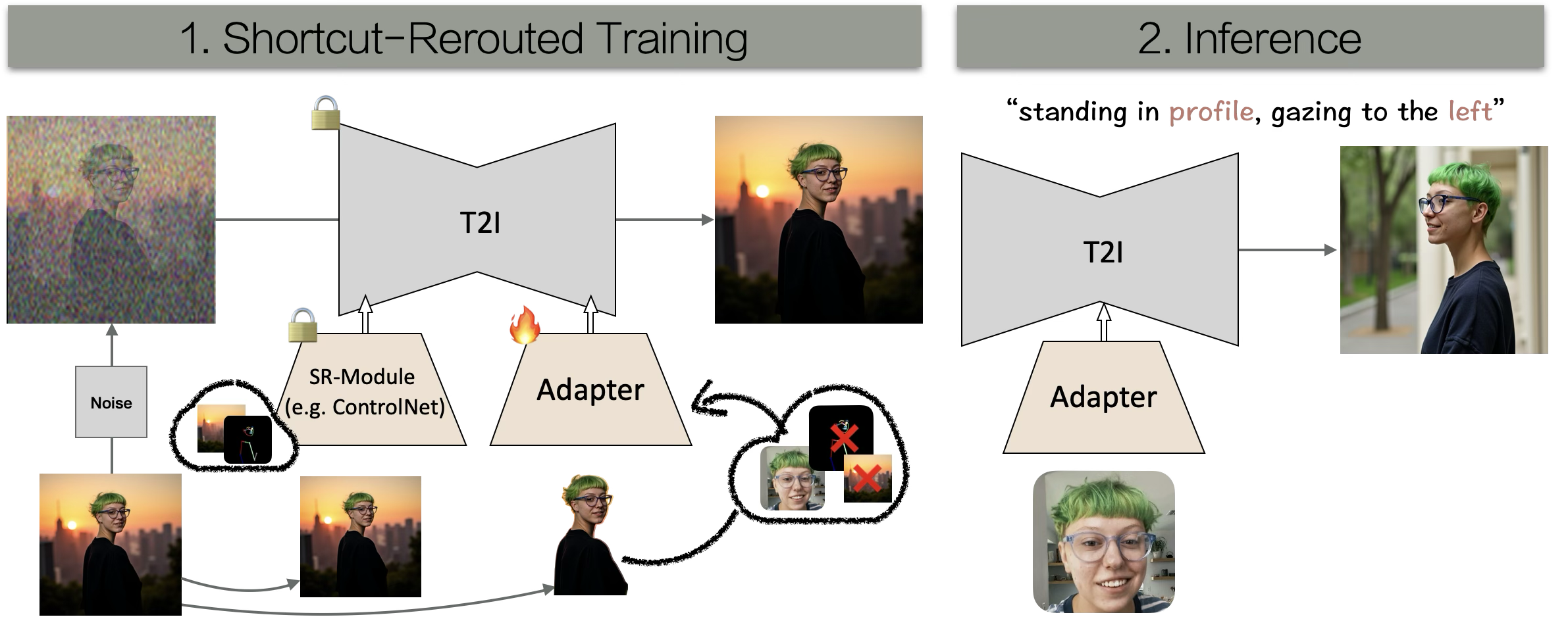}
    \caption{\textbf{Method.} The Shortcut-Rerouting (SR)-Module serves as a generic shortcut adapter that can take various forms—such as a ControlNet, LoRA, or IP-Adapter—depending on the confounding factor being addressed (e.g., pose, distribution, or style). Illustrated here is the case of SR with ControlNet, where pose and expression cues are explicitly rerouted via the ControlNet during adapter training. At inference time, the ControlNet is removed, restoring independent pose and expression control from the text prompt alone.
    } 
    \label{fig:method}
\end{figure}

We propose \textbf{Shortcut-Rerouted (SR) Adapter Training}. As demonstrated in \cref{fig:method}, our key insight is to reroute the shortcuts, i.e. the influence of $C$, through an auxiliary module $\mathcal{S}_C$, which explicitly utilizes the confounding factors. The generative process is modified to:
\begin{equation}\label{eq:sr-gen}
\hat{X} = G(\mathcal{A}(X), \mathcal{S}_C(C)),
\end{equation}

where $\mathcal{S}_C$ is a pre-trained and frozen module that directly provides $C$ to the generator, i.e. establishes the \emph{shortcuts}. The revised training objective becomes:
\begin{equation}\label{eq:sr-obj}
\mathbb{E}_{X}\big[\mathcal{L}\big(G(\mathcal{A}(X), \mathcal{S}_C(C)), X\big)\big],
\end{equation}

which ensures that the confounding factors are explained away by $\mathcal{S}_C$, leaving $\mathcal{A}(X)$ with no incentive to encode them. In effect, we turn the entanglement problem into a modular decomposition: $\mathcal{A}$ is pressured to specialize in $T$, while $\mathcal{S}_C$ accounts for $C$ during training.

Finally, \textbf{the shortcut module $\mathcal{S}_C$ is removed during inference}, recovering the original model, but equipped with a disentangled adapter. Now, the generative process in inference, $\hat{X} = G(\mathcal{A}(X))$, is less likely be impacted by the confounding factors from $X$.

This formulation reflects a general principle: by \emph{explicitly modeling} nuisance factors during training, we prevent the adapter from internalizing them, yielding cleaner and more robust representations.

\subsection{Personalized Generation via Shortcut-Rerouted(SR) T2I Adapter Training}

\begin{wrapfigure}{r}{0.4\linewidth}
    \vspace{-1em}
    \centering\includegraphics[width=\linewidth]{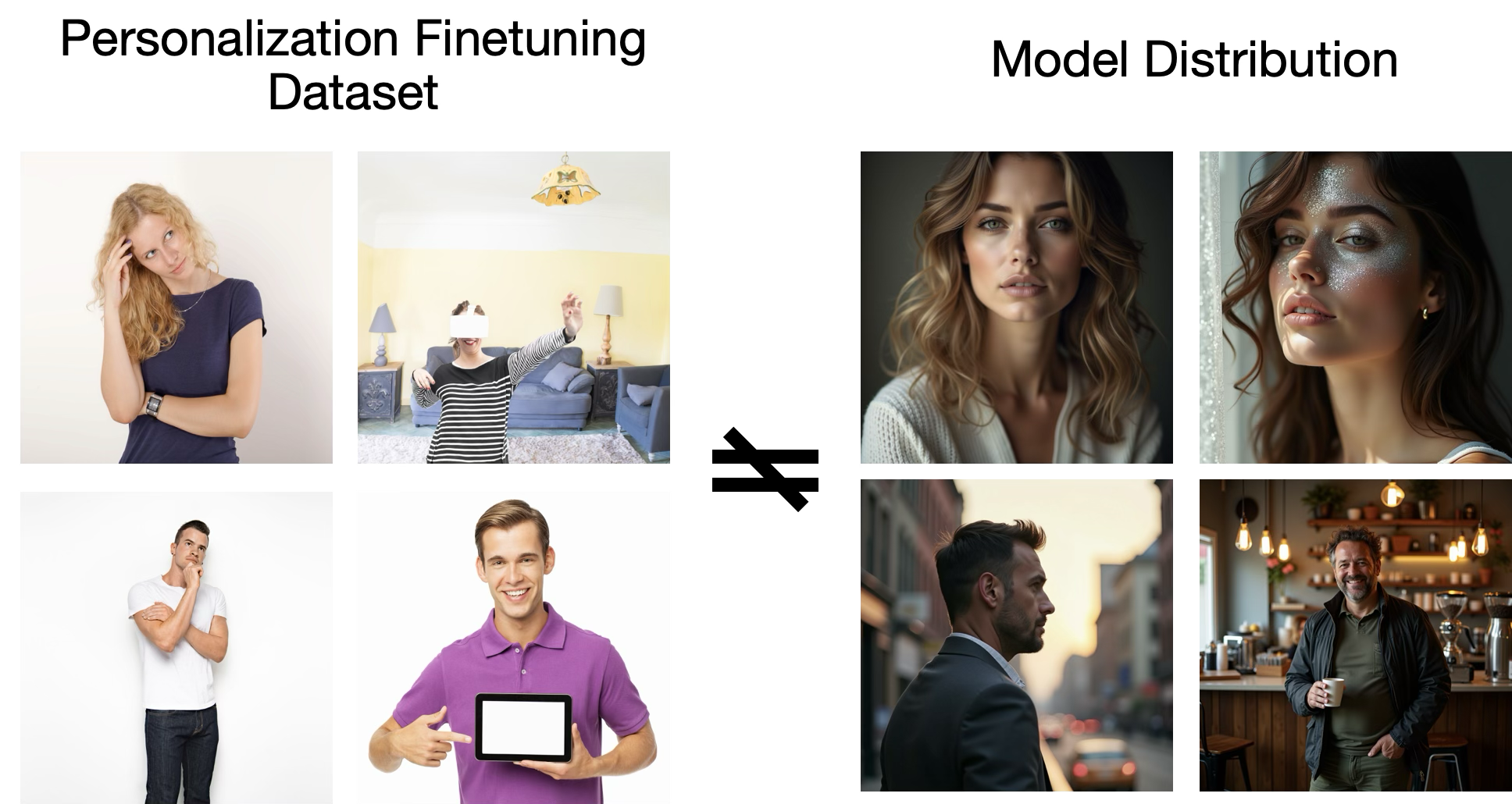}
    \caption{\textbf{Distribution shift between model and finetuning dataset.} Due to the distribution shift, directly training a personalization adapter on the finetuning dataset leads to degraded quality.}
    \label{fig:enter-label}
\end{wrapfigure}

Adapter training aims to steer a frozen text-to-image model by injecting additional signals—typically derived from a reference image—into its generation process. In the setting of personalized generation, adapters are lightweight modules that encode the input subject's identity and modulate the diffusion model to personalize its output accordingly.

\subsubsection {Instantiations of SR Module}
A central challenge in conventional adapter training is that the adapter often encodes confounding factors—such as distribution biases in the fine-tuning dataset or pose and lighting leakage from the input images—thereby entangling the target identity with spurious features.
The SR module $\mathcal{S}_C$ in \cref{eq:sr-gen} is versatile, and can be realized by many different modules capable of absorbing specific confounders.
While in principle multiple such modules can be composed to form a single unified SR module, in this work we focus on two primary instantiations: SR-LoRA, which addresses dataset-level distribution shifts, and SR-CN, which handles pose and expression leakage. 

% G-LoRA
\subsubsection{SR-LoRA: Addressing distribution shift}
The first application of Shortcut-Rerouted Adapter Training addresses the issue of \emph{distribution shift} between the model distribution and the data distribution used during adapter finetuning. In many real-world scenarios—particularly with proprietary models such as Flux—the training distribution of the backbone model is unknown or opaque. Meanwhile, personalization pipelines often finetune on curated datasets with specific styles, subjects, or domains. This mismatch introduces a latent confounding factor: \textit{the domain gap between the foundation model and the finetuning data}.

To absorb this domain-induced shortcut, we instantiate $\mathcal{S}_C$ as a light-weight LoRA module, trained specifically to capture this distributional gap. Concretely, we pretrain this LoRA on the finetuning dataset (e.g., studio-lit identity images), allowing it to absorb the dataset-specific style, lighting, and low-level features that differ from the base model's prior. During adapter training, we then freeze the LoRA and train the identity encoder $\mathcal{A}$ as the only active module, allowing it to focus solely on identity, independent of the dataset domain:
$$
\hat{X} = G(\mathcal{A}(X), \mathcal{S}_C(C)),
$$
where $\mathcal{S}_C$ provides the latent adjustment required to bridge the domain gap, rerouting the shortcut through a controlled path. As in our general formulation, this ensures that $\mathcal{A}(X)$ is no longer incentivized to account for the domain discrepancy, and instead focuses on learning a representation faithful to $T$.

\begin{figure}[!t]
    \centering
\includegraphics[width=1.0\linewidth]{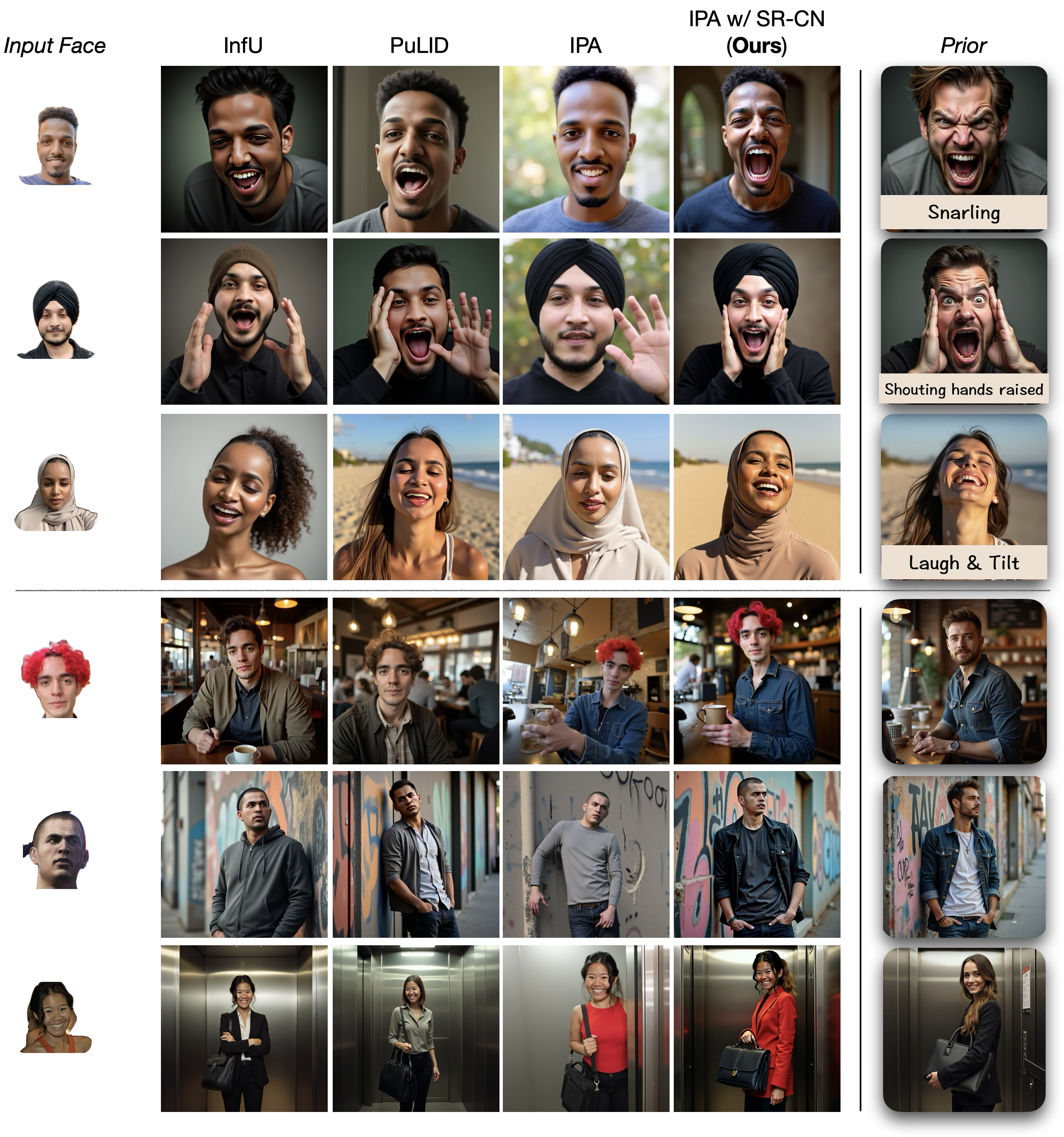}
    \caption{\textbf{Qualitative comparison of different ``face'' adapters.} \emph{Top}: close-up portraits with varied expressions. \emph{Bottom}: full-body generations. Our approach preserves the visual prior more faithfully, enabling expressive and identity-consistent personalized image generation. }
    \label{fig:qual-face}
\end{figure}

At inference time, we remove $\mathcal{S}_C$, resulting in a generation pipeline governed solely by $\mathcal{A}(X)$. This yields identity adapters that generalize beyond the specific visual domain of the training data and respond more reliably to test-time prompts across domains. In effect, this use case demonstrates that even abstract or latent confounders—such as dataset shift—can be systematically absorbed via shortcut modules, extending our methodology beyond structured confounds like pose or expression.

\subsubsection{SR-CN: Addressing Pose and Expression Leakage} 

Going beyond absorbing distribution shift, we aim to address the challenge of absorbing the shortcuts of expression and pose from the input image during inference.
The target factor $T$ is facial identity, while the target confounding factors $C_{CN}$ for this SR module include head pose and facial expression. To absorb $C_{CN}$, we employ a pre-trained ControlNet~\citep{ControlNet} module $\mathcal{S}_{CN}$ that conditions generation on pose and expression maps derived from the training images.

During adapter training, we augment the generative pipeline as follows: given a training image $X$ and its corresponding identity $T$ and pose/expression $C_{CN}$, we generate pose/expression maps (e.g., via pose estimation and landmark detection) and feed these to $\mathcal{S}_{CN}$. The adapter $\mathcal{A}$ is then trained to inject identity alone, while $\mathcal{S}_{CN}$ accounts for the pose and expression. The overall objective in adapter training stage becomes:
\begin{equation}\label{eq:sr-lora-cn-loss}
\mathcal{L}\big(G(\mathcal{A}(T), \mathcal{S}_{CN}(C_{CN})), X\big)
\end{equation}
which, as shown in \cref{fig:overview-of-results}, leads to adapters that are robust across a wide range of poses and expressions. See~\cref{fig:method} for an illustration.

This approach highlights the generality and modularity of our method: by providing a controlled pathway for spurious factors, we relieve the adapter from modeling them, resulting in cleaner and more generalizable representations.

\section{Experiments}
\label{sec:exp}

In this section, we show a number of experiments for different instance of Shortcut-Rerouted Adapter Training.  First, we show results for Shortcut Rerouting in the setting of `face' adapters using both LoRA and ControlNet as Shortcut Rerouting mechanisms (\cref{sec:exp:face}). The resulting adapters demonstrate improved prior preservation, head pose control and expression control.  Then, we show results for Shortcut Rerouting in the setting of `body' adapters (\cref{sec:exp:sr-cn-body}).

\subsection{Experimental Setup}

\textbf{\textit{Datasets.}} We curate an internal large-scale dataset of a few million high-quality human images, filtered to retain only single-subject photos and remove low-quality, NSFW, or watermarked content. To accelerate training, we bucket images by aspect ratio and cache auxiliary modalities such as landmarks, segmentation masks, and text embeddings.
For adapter inputs, we extract and align face crops using facial landmarks; for full-body, we extract body crops via segmentation and apply background removal. Captions are generated using Qwen2.5-14B (text) and InternViT-300M-V2.5 (vision), both state-of-the-art large-scale captioning models. We also provide the details and visualization for test input images and prompts in the Appendix.

\textbf{\textit{Training and Implementation details.}}
All methods are implemented in \texttt{PyTorch}~\citep{paszke2019pytorch} using the \texttt{HuggingFace Diffusers}~\citep{von-platen-etal-2022-diffusers} framework, based on the \texttt{FLUX.1 [Dev]}~\citep{FLUX} model with a DiT~\citep{Peebles2023DiT} backbone and Conditional Flow Matching objective~\citep{EsserKSD3}. Training is performed on 8$\times$A100 GPUs (80GB each) using AdamW~\citep{AdamW} with a learning rate of $5\mathrm{e}{-5}$ and a global batch size of 32 for 250K iterations. Inference is standardized across all methods with IP scale 1.0, CFG 3.5, 28 steps, and $1024 \times 1024$ resolution. For identity encoding, we use \texttt{openai/clip-vit-large-patch14}~\citep{CLIP}.

\textbf{\textit{Metrics}} used to measure id preservation, prompt following, and prior preservation are follows:

\begin{enumerate}[leftmargin=*]
    \item \textbf{FaceNet Id. $(\uparrow)$} -- cosine similarity of the FaceNet~\citep{schroff2015facenet} embeddings of the generated image and the input subject. 
    \item \textbf{LLM Id. $(\uparrow)$} -- a LLM-as-a-judge score for holistic identity. A good metric for personalization should capture the resemblance of the face, head, and hair. While existing Face metrics based on recognition models can capture the cropped face, it does not measure the head/hair.  To have a more holistic measure, we use LLM-as-a-judge similar to recent studies~\citep{luo2024stylus}. 
    \item \textbf{LLM Expr. $(\uparrow)$} -- a LLM-as-a-judge score for alignment between the expression specification in the prompt, and the expression in the generated images.
     \item \textbf{EMOCA Sim. $(\uparrow)$} -- cosine similarity between facial expression embeddings of the generated image and the prior image. The embeddings are extracted using EMOCA model \citep{danvevcek2022emoca}.
    
    \item \textbf{Head Pose. $(\downarrow)$} -- the mean absolute difference in head orientation (i.e. yaw, pitch, and roll) between the generated and prior images, measured in degrees. We use HopeNet~\citep{ruiz2018fine} to estimate these angles.
    \item \textbf{Body Pose. $(\downarrow)$} -- the mean L2 distance between estimated 2D body keypoints of the generated and prior images in pixel space. HRNet~\citep{sun2019hrnet} is used for keypoint estimation.
    \item \textbf{Prior (LPIPS) $(\downarrow)$} -- the LPIPS~\citep{zhang2018perceptual} between the generated image and the prior image.  Used to measure how much the generated image deviated from the prior.
\end{enumerate}
More details such as the instructions for the LLM-as-a-judge metrics can be found in the Appendix.

\textbf{\textit{Baselines.}}
Our experiments are conducted under two different settings: one where only the \textit{face} is used as input, and another where the fully \textit{body} is provided. Baselines for face-input setting include InfU~\citep{jiang2025infiniteyou}, PuLID~\citep{PuLID}, and an IP Adapter~\citep{ye2023ipadapter} trained by us without shortcut rerouting. For the body-input setting, we include an open source model from InstantX, namely \texttt{InstantX/FLUX.1-dev-IP-Adapter}~\citep{flux-ipa}. This is a general-purpose IPA and not specifically trained for human faces, but serves as a representative baseline for comprehensive evaluation.

\subsection{``Face'' Adapters}
\label{sec:exp:face}

\begin{table}[t]
\centering
\caption{\textbf{Quantitative comparison for ``face'' adapters.} Our Shortcut-Rerouted methods (SR-LoRA and SR-ControlNet) outperform prior baselines in head pose control and prior preservation, while maintaining competitive identity fidelity. All models are based on Flux Dev. }
\label{tab:face-selected}
\small
\resizebox{\linewidth}{!}{%
\begin{tabular}{lcccccc}
\toprule
\textbf{Method} & \textbf{LLM Id. $\uparrow$} & \textbf{FaceNet Id. $\uparrow$} & \textbf{LLM Expr. $\uparrow$} & \textbf{EMOCA Expr. $\uparrow$} & \textbf{Head Pose $\downarrow$} & \textbf{Prior (LPIPS) $\downarrow$} \\
\midrule
InfU~\citep{jiang2025infiniteyou}                               & 3.3824 & \uline{0.7402} & \textbf{3.7664} &0.5420 & 17.7139 & 0.4490 \\
PuLID~\citep{PuLID}                           & 4.2826 & \textbf{0.7742} & 3.5899 & 0.4890 & 17.5345 & 0.4584 \\
IPA~\citep{ye2023ipadapter}                  & \uline{4.7929}& 0.7150 & 3.0714 & 0.3470 & 16.1199 & 0.4800 \\
SR-LoRA IPA (\textbf{Ours})       & 4.7194 & 0.6708 & 3.4286 & 0.4580 & \uline{13.2701} & \uline{0.4330} \\
SR-CN IPA (\textbf{Ours}) & \textbf{4.7941}& 0.7118& \uline{3.6934} & \textbf{0.5800} & \textbf{12.6755}& \textbf{0.3937}\\
\bottomrule
\end{tabular}
}
\end{table}

\begin{table}[t]
\centering
\caption{\textbf{Quantitative evaluation of ``body'' personalization methods.}
% We compare personalization adapters on identity fidelity (LLM-based and FaceNet-based), expression control, head and body pose accuracy, and prior preservation (LPIPS).
Our SR-ControlNet outperforms both InstantX and baseline IPA across all metrics, showing improved disentanglement and better adherence to pose and expression prompts without sacrificing identity.}
\label{tab:sr-cn-body}
\small
\resizebox{\linewidth}{!}{%
\begin{tabular}{lccccccc}
    \toprule
    \textbf{Method} & \textbf{LLM Id. $\uparrow$} & \textbf{FaceNet Id. $\uparrow$} & \textbf{LLM Expr. $\uparrow$} & \textbf{EMOCA Expr. $\uparrow$} & \textbf{Head Pose $\downarrow$} & \textbf{Body Pose $\downarrow$} & \textbf{Prior (LPIPS) $\downarrow$} \\
    \midrule
    InstantX~\citep{instantx2024flux} & 2.9930 & 0.3533 & \uline{3.4736}& 0.4687 & 25.97 & 186.7454 & 0.5075 \\
    IPA~\citep{ye2023ipadapter} & \uline{4.5986} & \uline{0.5733} & 3.3000 & 0.3466 & \uline{20.70} & \uline{167.4000} & \uline{0.4566} \\
    SR-CN IPA (\textbf{Ours}) & \textbf{4.6510} &\textbf{ 0.5857} & \textbf{3.5263} & \textbf{0.4794} &\textbf{ 18.05 }& \textbf{137.6888} & \textbf{0.4133} \\
    \bottomrule
\end{tabular}
}
\end{table}

\textit{Absorbing distribution shift.}
Our first key result, shown in~\cref{tab:face-selected}, is the substantial improvement in prior preservation scores for both SR-LoRA and SR-CN. These gains are also clearly reflected in the qualitative examples in~\cref{fig:qual-face}, where the image layout, texture, and overall visual quality remain closely aligned with the reference prior. In contrast, all baseline methods—including the IPA variant without shortcut rerouting—exhibit noticeable deviations in texture and scene fidelity. The improvement in visual consistency highlights the effectiveness of shortcut rerouting in absorbing distributional differences during training.

\textit{Enabling text-guided control of pose and expression.}
Our second key result is that SR-CN  better preserves and generalizes over mutable aspects of a person’s identity—such as pose and expression—compared to standard IPA. \textit{Vanilla IPA suffers from pose and expression shortcuts, often copying these attributes directly from the input image.} For example, in the first row of~\cref{fig:qual-face}, the subject is smiling in the reference image, causing the output to ignore the prompt-specified expression of “snarling.” Other baselines, such as InfU and PuLID, also struggle with identity fidelity, showing noticeable inaccuracies in head shape and hairstyle. It is worth noting that unfortunately face identity distance cannot measure this identity shift caused by head shape and hairstyle. In contrast, the more disentangled adapter trained with shortcut rerouting not only respects prompt-driven expression but also supports more natural and coherent full-body generations (see bottom half of~\cref{fig:qual-face}).

\subsection{``Body'' Adapters}
\label{sec:exp:sr-cn-body}

\begin{figure}
    \centering
    \includegraphics[width=0.9\linewidth]{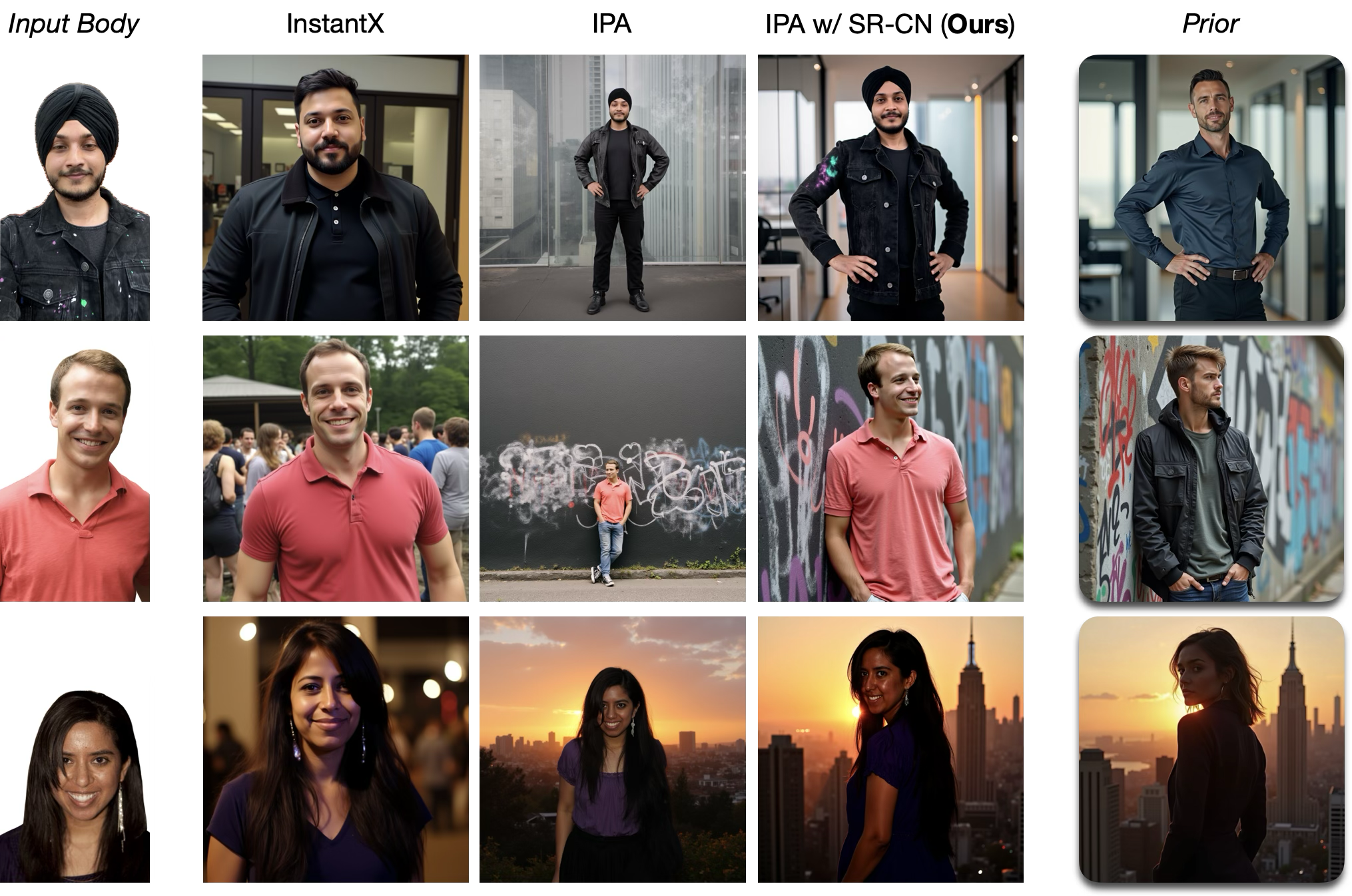}
    \caption{\textbf{Qualitative comparison of different ``body'' adapters.} Our approach shows much stronger identity preservation than InstantX~\citep{instantx2024flux}, and much better adherence to the prior and enhanced image quality when compared to vanilla IPA~\citep{ye2023ipadapter}.}
    \label{fig:qual-body}
\end{figure}

\textit{Beyond just face: Capturing identity aspects like body type, clothing, and limb proportions.}
We explore training adapters using full-body crops as input, which provide a richer signal for capturing holistic identity traits such as body type, clothing, and limb proportions—factors critical for realism and character consistency in downstream generations. \textit{However, this richer input also increases the risk of shortcut learning, particularly the tendency to copy the body pose from the reference image.} As a result, desired applications like reposing a subject through text prompts become more difficult.

As shown in~\cref{tab:sr-cn-body}, IPA trained with shortcut rerouting achieves the highest performance across all metrics: identity fidelity, pose controllability, and prior adherence. These improvements are clearly visible in~\cref{fig:qual-body}, where SR-CN IPA not only preserves subject identity more faithfully, but also \textit{produces outputs that are more consistent with the prior image layout and appearance}.

\subsection{``Background'' Adapter}

\begin{figure}[!t]
    \centering
    \includegraphics[width=\linewidth]{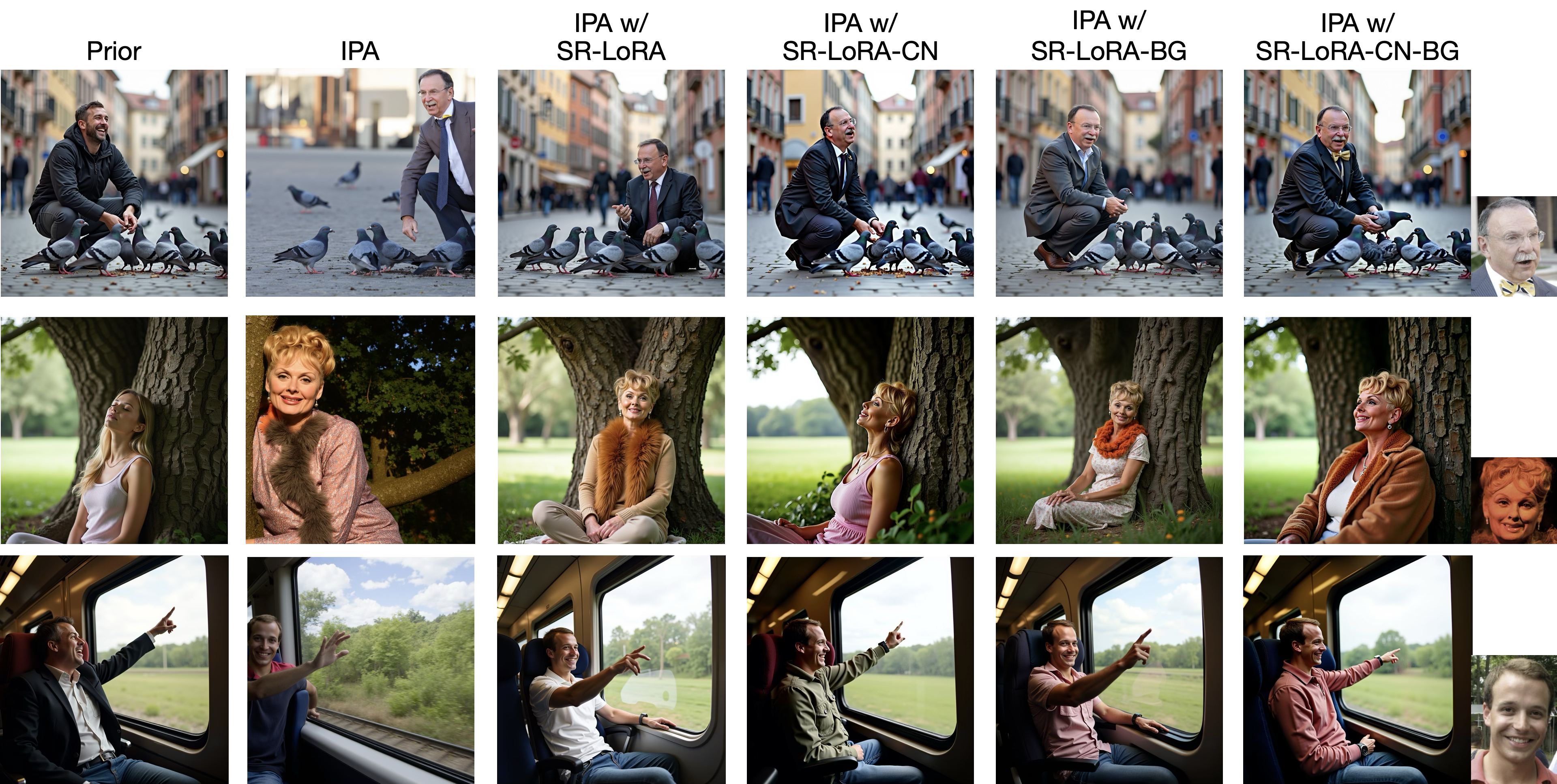}
    \caption{\textbf{SR-Training is a versatile framework supporting different combinations of shortcut modules}. The LoRA shortcut mitigates quality degradation, producing generations more consistent with the prior compared to the baseline. The ControlNet (CN) shortcut preserves pose priors, while the background (BG) shortcut prevents lighting leakage from the input. Notably, SR-LoRA-CN follows the pose of the prior but deviates in the background, whereas SR-LoRA-BG preserves the background but deviates slightly in pose. Finally, SR-LoRA-CN-BG aligns closely with both pose and background, thereby isolating and injecting only the target identity.}

    \label{fig:all-adapters}
    \vspace{-1em}
\end{figure}

Additional variants that include further modules (e.g., background adapters) and extended combinations such as SR-LoRA+CN, and SR-LoRA+CN+BG are discussed in the Appendix, along with ablation results illustrating their complementary effects.
As an example depicted in~\cref{fig:all-adapters}, the LoRA shortcut substantially alleviates quality degradation, the ControlNet (CN) shortcut reliably maintains pose priors, and the background (BG) shortcut effectively suppresses illumination leakage.

\textbf{Limitations.}
In this work, we introduced Shortcut-Rerouted Adapter Training, a simple yet broadly applicable framework for disentangling spurious correlations in text-to-image personalization.
One limitation with the current evaluation is our focus on encoder-based adapter training.  In principle, our approach could also be applicable to LoRA training, such as learning a style LoRA free of undesired shifts like layout, or content.  
Specifically for the task of personalization, one limitation is the fact that we applied Shortcut-Rerouting to IP-Adapter only, which is a fairly simple baseline.  Applying our approach to stronger baselines might lead to overall better performance.

\textbf{Ethical Considerations}. As our method improves identity preservation and expression control in personalized generation, it naturally raises concerns about misuse, particularly in the creation of hyper-realistic synthetic identities or deepfakes. We acknowledge that enhanced controllability and realism may lower the barrier for malicious use. To mitigate such risks, we advocate for responsible deployment practices, such as model watermarking, usage restrictions, and alignment with ethical frameworks for generative media.

\section{Conclusion}

We introduced Shortcut-Rerouting, a simple yet general framework for disentangling spurious correlations in text-to-image personalization. By explicitly providing shortcut pathways for confounding factors during training—via modules such as LoRA or ControlNet—we prevent adapters from internalizing undesirable attributes like pose, expression, or domain-specific biases. This leads to cleaner, more controllable representations that preserve identity while restoring the model’s ability to respond to prompt-based instructions. Our experiments across both face and full-body personalization demonstrate improved controllability, prior preservation, and generation quality. 

More broadly, our results suggest a general principle: when training models to focus on what matters, it is often most effective to explicitly route away what does not.
We believe this perspective of Shortcut Rerouting has implications beyond personalization: the idea of absorbing confounding variation through targeted pathways can inform future approaches to modular, interpretable, and more controllable generative systems.

\newpage
\bibliographystyle{plainnat}
\bibliography{bib}

\begin{thebibliography}{42}
\providecommand{\natexlab}[1]{#1}
\providecommand{\url}[1]{\texttt{#1}}
\expandafter\ifx\csname urlstyle\endcsname\relax
  \providecommand{\doi}[1]{doi: #1}\else
  \providecommand{\doi}{doi: \begingroup \urlstyle{rm}\Url}\fi

\bibitem[Arjovsky et~al.(2019)Arjovsky, Bottou, Gulrajani, and Lopez{-}Paz]{mart2019invariant_risk_mini}
Mart{\'{\i}}n Arjovsky, L{\'{e}}on Bottou, Ishaan Gulrajani, and David Lopez{-}Paz.
\newblock Invariant risk minimization.
\newblock \emph{CoRR}, abs/1907.02893, 2019.

\bibitem[Dan{\v{e}}{\v{c}}ek et~al.(2022)Dan{\v{e}}{\v{c}}ek, Black, and Bolkart]{danvevcek2022emoca}
Radek Dan{\v{e}}{\v{c}}ek, Michael~J Black, and Timo Bolkart.
\newblock Emoca: Emotion driven monocular face capture and animation.
\newblock In \emph{Proceedings of the IEEE/CVF Conference on Computer Vision and Pattern Recognition}, pages 20311--20322, 2022.

\bibitem[Deng et~al.(2022)Deng, Guo, Yang, Xue, Kotsia, and Zafeiriou]{ArcFace}
Jiankang Deng, Jia Guo, Jing Yang, Niannan Xue, Irene Kotsia, and Stefanos Zafeiriou.
\newblock {ArcFace}: Additive angular margin loss for deep face recognition.
\newblock \emph{{IEEE} Transactions on Pattern Analysis and Machine Intelligence}, 44\penalty0 (10):\penalty0 5962--5979, oct 2022.
\newblock \doi{10.1109/tpami.2021.3087709}.
\newblock URL \url{https://doi.org/10.1109%2Ftpami.2021.3087709}.

\bibitem[Esser et~al.(2024)Esser, Kulal, Blattmann, Entezari, M{\"{u}}ller, Saini, Levi, Lorenz, Sauer, Boesel, Podell, Dockhorn, English, and Rombach]{EsserKSD3}
Patrick Esser, Sumith Kulal, Andreas Blattmann, Rahim Entezari, Jonas M{\"{u}}ller, Harry Saini, Yam Levi, Dominik Lorenz, Axel Sauer, Frederic Boesel, Dustin Podell, Tim Dockhorn, Zion English, and Robin Rombach.
\newblock Scaling rectified flow transformers for high-resolution image synthesis.
\newblock In \emph{Proceedings of the International Conference on Machine Learning (ICML)}. OpenReview.net, 2024.

\bibitem[Gal et~al.(2021)Gal, Patashnik, Maron, Chechik, and Cohen-Or]{gal2021stylegan}
Rinon Gal, Or~Patashnik, Haggai Maron, Gal Chechik, and Daniel Cohen-Or.
\newblock Stylegan-nada: Clip-guided domain adaptation of image generators.
\newblock \emph{arXiv preprint arXiv:2108.00946}, 2021.

\bibitem[Gal et~al.(2022)Gal, Alaluf, Atzmon, Patashnik, Bermano, Chechik, and Cohen-Or]{gal2022image}
Rinon Gal, Yuval Alaluf, Yuval Atzmon, Or~Patashnik, Amit~H Bermano, Gal Chechik, and Daniel Cohen-Or.
\newblock An image is worth one word: Personalizing text-to-image generation using textual inversion.
\newblock \emph{arXiv preprint arXiv:2208.01618}, 2022.

\bibitem[Geirhos et~al.(2019)Geirhos, Rubisch, Michaelis, Bethge, Wichmann, and Brendel]{Geirhos19texure_bias}
Robert Geirhos, Patricia Rubisch, Claudio Michaelis, Matthias Bethge, Felix~A. Wichmann, and Wieland Brendel.
\newblock Imagenet-trained cnns are biased towards texture; increasing shape bias improves accuracy and robustness.
\newblock In \emph{ICLR}. OpenReview.net, 2019.

\bibitem[Geirhos et~al.(2020)Geirhos, Jacobsen, Michaelis, Zemel, Brendel, Bethge, and Wichmann]{geirhos2020shortcut}
Robert Geirhos, J{\"{o}}rn{-}Henrik Jacobsen, Claudio Michaelis, Richard~S. Zemel, Wieland Brendel, Matthias Bethge, and Felix~A. Wichmann.
\newblock Shortcut learning in deep neural networks.
\newblock \emph{Nat. Mach. Intell.}, 2\penalty0 (11):\penalty0 665--673, 2020.

\bibitem[Guo et~al.(2023)Guo, Yang, Rao, Liang, Wang, Qiao, Agrawala, Lin, and Dai]{guo2023animatediff}
Yuwei Guo, Ceyuan Yang, Anyi Rao, Zhengyang Liang, Yaohui Wang, Yu~Qiao, Maneesh Agrawala, Dahua Lin, and Bo~Dai.
\newblock Animatediff: Animate your personalized text-to-image diffusion models without specific tuning.
\newblock \emph{arXiv preprint arXiv:2307.04725}, 2023.

\bibitem[Guo et~al.(2024)Guo, Wu, Chen, Chen, and He]{PuLID}
Zinan Guo, Yanze Wu, Zhuowei Chen, Lang Chen, and Qian He.
\newblock Pulid: Pure and lightning {ID} customization via contrastive alignment.
\newblock \emph{CoRR}, abs/2404.16022, 2024.

\bibitem[Ho et~al.(2020)Ho, Jain, and Abbeel]{ho2020denoising}
Jonathan Ho, Ajay Jain, and Pieter Abbeel.
\newblock Denoising diffusion probabilistic models.
\newblock \emph{Advances in neural information processing systems}, 33:\penalty0 6840--6851, 2020.

\bibitem[Hu et~al.(2022)Hu, Shen, Wallis, Allen-Zhu, Li, Wang, Wang, Chen, et~al.]{hu2022lora}
Edward~J Hu, Yelong Shen, Phillip Wallis, Zeyuan Allen-Zhu, Yuanzhi Li, Shean Wang, Lu~Wang, Weizhu Chen, et~al.
\newblock Lora: Low-rank adaptation of large language models.
\newblock \emph{ICLR}, 1\penalty0 (2):\penalty0 3, 2022.

\bibitem[InstantX(2024)]{instantx2024flux}
InstantX.
\newblock {FLUX.1-dev-IP-Adapter}.
\newblock \url{https://huggingface.co/InstantX/FLUX.1-dev-IP-Adapter}, 2024.
\newblock Accessed: 2024-11-01.

\bibitem[Islam et~al.(2020)Islam, Jia, and Bruce]{Islam20PosInfo}
Md.~Amirul Islam, Sen Jia, and Neil D.~B. Bruce.
\newblock How much position information do convolutional neural networks encode?
\newblock In \emph{ICLR}. OpenReview.net, 2020.

\bibitem[Jiang et~al.(2025)Jiang, Yan, Jia, Liu, Kang, and Lu]{jiang2025infiniteyou}
Liming Jiang, Qing Yan, Yumin Jia, Zichuan Liu, Hao Kang, and Xin Lu.
\newblock Infiniteyou: Flexible photo recrafting while preserving your identity.
\newblock \emph{arXiv preprint arXiv:2503.16418}, 2025.

\bibitem[Jones et~al.(2024)Jones, Wang, Kumari, Bau, and Zhu]{jones2024customizing}
Maxwell Jones, Sheng-Yu Wang, Nupur Kumari, David Bau, and Jun-Yan Zhu.
\newblock Customizing text-to-image models with a single image pair.
\newblock In \emph{SIGGRAPH Asia 2024 Conference Papers}, pages 1--13, 2024.

\bibitem[Labs(2024)]{FLUX}
Black~Forest Labs.
\newblock Flux.
\newblock \url{https://github.com/black-forest-labs/flux}, 2024.

\bibitem[Loshchilov and Hutter(2019)]{AdamW}
Ilya Loshchilov and Frank Hutter.
\newblock Decoupled weight decay regularization.
\newblock In \emph{ICLR}, 2019.

\bibitem[Luo et~al.(2024)Luo, Wong, Trabucco, Huang, Gonzalez, Salakhutdinov, Stoica, et~al.]{luo2024stylus}
Michael Luo, Justin Wong, Brandon Trabucco, Yanping Huang, Joseph~E Gonzalez, Ruslan Salakhutdinov, Ion Stoica, et~al.
\newblock Stylus: Automatic adapter selection for diffusion models.
\newblock \emph{Advances in Neural Information Processing Systems}, 37:\penalty0 32888--32915, 2024.

\bibitem[Luo et~al.(2021)Luo, Wei, Wen, Yang, Xie, Xu, and Tian]{luo2021rectifying}
Xu~Luo, Longhui Wei, Liangjian Wen, Jinrong Yang, Lingxi Xie, Zenglin Xu, and Qi~Tian.
\newblock Rectifying the shortcut learning of background for few-shot learning.
\newblock \emph{Advances in Neural Information Processing Systems}, 34:\penalty0 13073--13085, 2021.

\bibitem[OpenAI(2024)]{ChatGPT4o}
OpenAI.
\newblock Gpt-4o technical report, 2024.
\newblock URL \url{https://openai.com/index/gpt-4o}.
\newblock Accessed: 2025-05-22.

\bibitem[Ostris(2024)]{ostrislora}
Ostris.
\newblock Flux.1-schnell-training-adapter, 2024.
\newblock URL \url{https://huggingface.co/ostris/FLUX.1-schnell-training-adapter}.
\newblock Accessed: 2025-10-15.

\bibitem[Paszke et~al.(2019)Paszke, Gross, Massa, Lerer, Bradbury, Chanan, Killeen, Lin, Gimelshein, Antiga, et~al.]{paszke2019pytorch}
Adam Paszke, Sam Gross, Francisco Massa, Adam Lerer, James Bradbury, Gregory Chanan, Trevor Killeen, Zeming Lin, Natalia Gimelshein, Luca Antiga, et~al.
\newblock Pytorch: An imperative style, high-performance deep learning library.
\newblock \emph{Advances in neural information processing systems}, 32, 2019.

\bibitem[Peebles and Xie(2023)]{Peebles2023DiT}
William Peebles and Saining Xie.
\newblock Scalable diffusion models with transformers.
\newblock In \emph{Proceedings of the IEEE/CVF International Conference on Computer Vision (ICCV)}, pages 4172--4182. {IEEE}, 2023.

\bibitem[Qian et~al.(2025{\natexlab{a}})Qian, Wang, Patashnik, Heravi, Ostashev, Tulyakov, Cohen-Or, and Aberman]{qian2025omni}
Guocheng Qian, Kuan-Chieh Wang, Or~Patashnik, Negin Heravi, Daniil Ostashev, Sergey Tulyakov, Daniel Cohen-Or, and Kfir Aberman.
\newblock Omni-id: Holistic identity representation designed for generative tasks.
\newblock In \emph{Proceedings of the Computer Vision and Pattern Recognition Conference}, pages 8786--8795, 2025{\natexlab{a}}.

\bibitem[Qian et~al.(2025{\natexlab{b}})Qian, Ostashev, Nemchinov, Assouline, Tulyakov, Wang, and Aberman]{qian2025composeme}
Guocheng~Gordon Qian, Daniil Ostashev, Egor Nemchinov, Avihay Assouline, Sergey Tulyakov, Kuan-Chieh~Jackson Wang, and Kfir Aberman.
\newblock Composeme: Attribute-specific image prompts for controllable human image generation.
\newblock \emph{arXiv preprint arXiv:2509.18092}, 2025{\natexlab{b}}.

\bibitem[Radford et~al.(2021)Radford, Kim, Hallacy, Ramesh, Goh, Agarwal, Sastry, Askell, Mishkin, Clark, et~al.]{CLIP}
Alec Radford, Jong~Wook Kim, Chris Hallacy, Aditya Ramesh, Gabriel Goh, Sandhini Agarwal, Girish Sastry, Amanda Askell, Pamela Mishkin, Jack Clark, et~al.
\newblock Learning transferable visual models from natural language supervision.
\newblock In \emph{International conference on machine learning}, pages 8748--8763. PMLR, 2021.

\bibitem[Ramesh et~al.(2022)Ramesh, Dhariwal, Nichol, Chu, and Chen]{ramesh2022hierarchical}
Aditya Ramesh, Prafulla Dhariwal, Alex Nichol, Casey Chu, and Mark Chen.
\newblock Hierarchical text-conditional image generation with clip latents.
\newblock \emph{arXiv preprint arXiv:2204.06125}, 1\penalty0 (2):\penalty0 3, 2022.

\bibitem[Rombach et~al.(2022)Rombach, Blattmann, Lorenz, Esser, and Ommer]{ldm}
Robin Rombach, Andreas Blattmann, Dominik Lorenz, Patrick Esser, and Bj{\"o}rn Ommer.
\newblock High-resolution image synthesis with latent diffusion models.
\newblock In \emph{CVPR}, pages 10684--10695, 2022.

\bibitem[Ruiz et~al.(2018)Ruiz, Chong, and Rehg]{ruiz2018fine}
Nataniel Ruiz, Eunji Chong, and James~M Rehg.
\newblock Fine-grained head pose estimation without keypoints.
\newblock In \emph{Proceedings of the IEEE Conference on Computer Vision and Pattern Recognition Workshops (CVPRW)}, 2018.

\bibitem[Ruiz et~al.(2023)Ruiz, Li, Jampani, Pritch, Rubinstein, and Aberman]{ruiz2023dreambooth}
Nataniel Ruiz, Yuanzhen Li, Varun Jampani, Yael Pritch, Michael Rubinstein, and Kfir Aberman.
\newblock Dreambooth: Fine tuning text-to-image diffusion models for subject-driven generation.
\newblock In \emph{CVPR}, pages 22500--22510, 2023.

\bibitem[Schroff et~al.(2015)Schroff, Kalenichenko, and Philbin]{schroff2015facenet}
Florian Schroff, Dmitry Kalenichenko, and James Philbin.
\newblock Facenet: A unified embedding for face recognition and clustering.
\newblock In \emph{Proceedings of the IEEE conference on computer vision and pattern recognition (CVPR)}, pages 815--823, 2015.

\bibitem[Sun et~al.(2019)Sun, Xiao, Liu, and Wang]{sun2019hrnet}
Ke~Sun, Bin Xiao, Dong Liu, and Jingdong Wang.
\newblock Deep high-resolution representation learning for human pose estimation.
\newblock In \emph{Proceedings of the IEEE Conference on Computer Vision and Pattern Recognition (CVPR)}, pages 5693--5703, 2019.

\bibitem[Team(2024)]{flux-ipa}
InstantX Team.
\newblock Instantx flux.1-dev ip-adapter page.
\newblock \url{https://huggingface.co/InstantX/FLUX.1-dev-IP-Adapter}, 2024.

\bibitem[von Platen et~al.(2022)von Platen, Patil, Lozhkov, Cuenca, Lambert, Rasul, Davaadorj, and Wolf]{von-platen-etal-2022-diffusers}
Patrick von Platen, Suraj Patil, Anton Lozhkov, Pedro Cuenca, Nathan Lambert, Kashif Rasul, Mishig Davaadorj, and Thomas Wolf.
\newblock Diffusers: State-of-the-art diffusion models.
\newblock \url{https://github.com/huggingface/diffusers}, 2022.

\bibitem[Voynov et~al.(2023)Voynov, Chu, Cohen-Or, and Aberman]{voynov2023p+}
Andrey Voynov, Qinghao Chu, Daniel Cohen-Or, and Kfir Aberman.
\newblock $ p+ $: Extended textual conditioning in text-to-image generation.
\newblock \emph{arXiv preprint arXiv:2303.09522}, 2023.

\bibitem[Wang et~al.(2024)Wang, Ostashev, Fang, Tulyakov, and Aberman]{wang2024moa}
Kuan-Chieh Wang, Daniil Ostashev, Yuwei Fang, Sergey Tulyakov, and Kfir Aberman.
\newblock Moa: Mixture-of-attention for subject-context disentanglement in personalized image generation.
\newblock In \emph{SIGGRAPH Asia 2024 Conference Papers}, pages 1--12, 2024.

\bibitem[Wang et~al.(2023)Wang, Wang, Xie, Qi, Shan, Wang, and Luo]{Wang2023StyleAdapterAS}
Zhouxia Wang, Xintao Wang, Liangbin Xie, Zhongang Qi, Ying Shan, Wenping Wang, and Ping Luo.
\newblock Styleadapter: A single-pass lora-free model for stylized image generation.
\newblock \emph{ArXiv}, abs/2309.01770, 2023.
\newblock URL \url{https://api.semanticscholar.org/CorpusID:261531689}.

\bibitem[Xiao et~al.(2023)Xiao, Yin, Freeman, Durand, and Han]{fastcomposer}
Guangxuan Xiao, Tianwei Yin, William~T Freeman, Fr{\'e}do Durand, and Song Han.
\newblock Fastcomposer: Tuning-free multi-subject image generation with localized attention.
\newblock \emph{arXiv preprint arXiv:2305.10431}, 2023.

\bibitem[Ye et~al.(2023)Ye, Zhang, Liu, Han, and Yang]{ye2023ipadapter}
Hu~Ye, Jun Zhang, Sibo Liu, Xiao Han, and Wei Yang.
\newblock Ip-adapter: Text compatible image prompt adapter for text-to-image diffusion models.
\newblock \emph{arXiv preprint arXiv:2308.06721}, 2023.

\bibitem[Zhang et~al.(2023)Zhang, Rao, and Agrawala]{ControlNet}
Lvmin Zhang, Anyi Rao, and Maneesh Agrawala.
\newblock Adding conditional control to text-to-image diffusion models, 2023.

\bibitem[Zhang et~al.(2018)Zhang, Isola, Efros, Shechtman, and Wang]{zhang2018perceptual}
Richard Zhang, Phillip Isola, Alexei~A Efros, Eli Shechtman, and Oliver Wang.
\newblock The unreasonable effectiveness of deep features as a perceptual metric.
\newblock In \emph{CVPR}, 2018.

\end{thebibliography}

%%%%%%%%%%%%%%%%%%%%%%%%%%%%%%%%%%%%%%%%%%%%%%%%%%%%%%%%%%%%
\newpage

\section{Checklist}
\begin{enumerate}

\item {\bf Claims}
    \item[] Question: Do the main claims made in the abstract and introduction accurately reflect the paper's contributions and scope?
    \item[] Answer: \answerYes{} % Replace by \answerYes{}, \answerNo{}, or \answerNA{}.
    \item[] Justification:
    % The introduction clearly outlines the motivation, i.e. to improve the personalization as well as edit-ability of adapters, and to provide guardrails in the learning process in order to prevent them from overfitting to incidental non-personalization attributes. We demonstrate the capability using two well-established methods. Quantitatively, our full-body adapter outperforms the baseline on all the four metrics and face-adapter outperforms by a wide margin on the head pose and prior preservation (demonstrating the ability to control pose of the subject from text ) while showing comparable results ont eh method outperforms the baselines on all Our method outperforms the 

\item {\bf Limitations}
    \item[] Question: Does the paper discuss the limitations of the work performed by the authors?
    \item[] Answer: \answerYes{} % Replace by \answerYes{}, \answerNo{}, or \answerNA{}.
    \item[] Justification: 

\item {\bf Theory assumptions and proofs}
    \item[] Question: For each theoretical result, does the paper provide the full set of assumptions and a complete (and correct) proof?
    \item[] Answer: \answerNA{} % Replace by \answerYes{}, \answerNo{}, or \answerNA{}.
    \item[] Justification: 

    \item {\bf Experimental result reproducibility}
    \item[] Question: Does the paper fully disclose all the information needed to reproduce the main experimental results of the paper to the extent that it affects the main claims and/or conclusions of the paper (regardless of whether the code and data are provided or not)?
    \item[] Answer: \answerYes{} % Replace by \answerYes{}, \answerNo{}, or \answerNA{}.
    \item[] Justification: 

\item {\bf Open access to data and code}
    \item[] Question: Does the paper provide open access to the data and code, with sufficient instructions to faithfully reproduce the main experimental results, as described in supplemental material?
    \item[] Answer: \answerNA{} % Replace by \answerYes{}, \answerNo{}, or \answerNA{}.
    \item[] Justification: At the time of submission, we are unable to release the code or dataset due to legal and compliance constraints within our organization. We recognize the value of open access and reproducibility and are actively exploring the possibility of releasing portions of the code or evaluation tools pending internal review. We provide detailed descriptions of our methodology, datasets, and evaluation setup in the paper and appendix to support reproducibility in principle.

\item {\bf Experimental setting/details}
    \item[] Question: Does the paper specify all the training and test details (e.g., data splits, hyperparameters, how they were chosen, type of optimizer, etc.) necessary to understand the results?
    \item[] Answer: \answerYes{} % Replace by \answerYes{}, \answerNo{}, or \answerNA{}.
    \item[] Justification: 

\item {\bf Experiment statistical significance}
    \item[] Question: Does the paper report error bars suitably and correctly defined or other appropriate information about the statistical significance of the experiments?
    \item[] Answer: \answerYes{} % Replace by \answerYes{}, \answerNo{}, or \answerNA{}.
    \item[] Justification: 

\item {\bf Experiments compute resources}
    \item[] Question: For each experiment, does the paper provide sufficient information on the computer resources (type of compute workers, memory, time of execution) needed to reproduce the experiments?
    \item[] Answer: \answerYes{} % Replace by \answerYes{}, \answerNo{}, or \answerNA{}.
    \item[] Justification: 
    
\item {\bf Code of ethics}
    \item[] Question: Does the research conducted in the paper conform, in every respect, with the NeurIPS Code of Ethics \url{https://neurips.cc/public/EthicsGuidelines}?
    \item[] Answer: \answerYes{} % Replace by \answerYes{}, \answerNo{}, or \answerNA{}.
    \item[] Justification: 

\item {\bf Broader impacts}
    \item[] Question: Does the paper discuss both potential positive societal impacts and negative societal impacts of the work performed?
    \item[] Answer: \answerYes{} % Replace by \answerYes{}, \answerNo{}, or \answerNA{}.
    \item[] Justification: 
    
\item {\bf Safeguards}
    \item[] Question: Does the paper describe safeguards that have been put in place for responsible release of data or models that have a high risk for misuse (e.g., pretrained language models, image generators, or scraped datasets)?
    \item[] Answer: \answerNA{} % Replace by \answerYes{}, \answerNo{}, or \answerNA{}.
    \item[] Justification: 

\item {\bf Licenses for existing assets}
    \item[] Question: Are the creators or original owners of assets (e.g., code, data, models), used in the paper, properly credited and are the license and terms of use explicitly mentioned and properly respected?
    \item[] Answer: \answerYes{} % Replace by \answerYes{}, \answerNo{}, or \answerNA{}.
    \item[] Justification: 

\item {\bf New assets}
    \item[] Question: Are new assets introduced in the paper well documented and is the documentation provided alongside the assets?
    \item[] Answer: \answerNA{} % Replace by \answerYes{}, \answerNo{}, or \answerNA{}.
    \item[] Justification: 

\item {\bf Crowdsourcing and research with human subjects}
    \item[] Question: For crowdsourcing experiments and research with human subjects, does the paper include the full text of instructions given to participants and screenshots, if applicable, as well as details about compensation (if any)? 
    \item[] Answer: \answerNA{} % Replace by \answerYes{}, \answerNo{}, or \answerNA{}.
    \item[] Justification: 

\item {\bf Institutional review board (IRB) approvals or equivalent for research with human subjects}
    \item[] Question: Does the paper describe potential risks incurred by study participants, whether such risks were disclosed to the subjects, and whether Institutional Review Board (IRB) approvals (or an equivalent approval/review based on the requirements of your country or institution) were obtained?
    \item[] Answer: \answerNA{} % Replace by \answerYes{}, \answerNo{}, or \answerNA{}.
    \item[] Justification: 

\item {\bf Declaration of LLM usage}
    \item[] Question: Does the paper describe the usage of LLMs if it is an important, original, or non-standard component of the core methods in this research? Note that if the LLM is used only for writing, editing, or formatting purposes and does not impact the core methodology, scientific rigorousness, or originality of the research, declaration is not required.
    %this research? 
    \item[] Answer: \answerYes{} % Replace by \answerYes{}, \answerNo{}, or \answerNA{}.
    \item[] Justification: 

\end{enumerate}

\newpage
\appendix
\appendix

\section {Acknowledgments}
We thank Denis Bondarev, Ergeta Muca,  Bridget Briley-Snook, Jackie Fuhrman, Jonathan Solichin, Julia Krysko, and Svitlana Stern for their guidance, production support, reviewing paper drafts, and coordination across AR engineering and creative efforts that made this work possible.

\section {Comparing LLM-as-a-judge with FaceNet as a metric for Identity Preservation}

\begin{figure}[ht]
    \centering
    \includegraphics[width=1.0\linewidth]{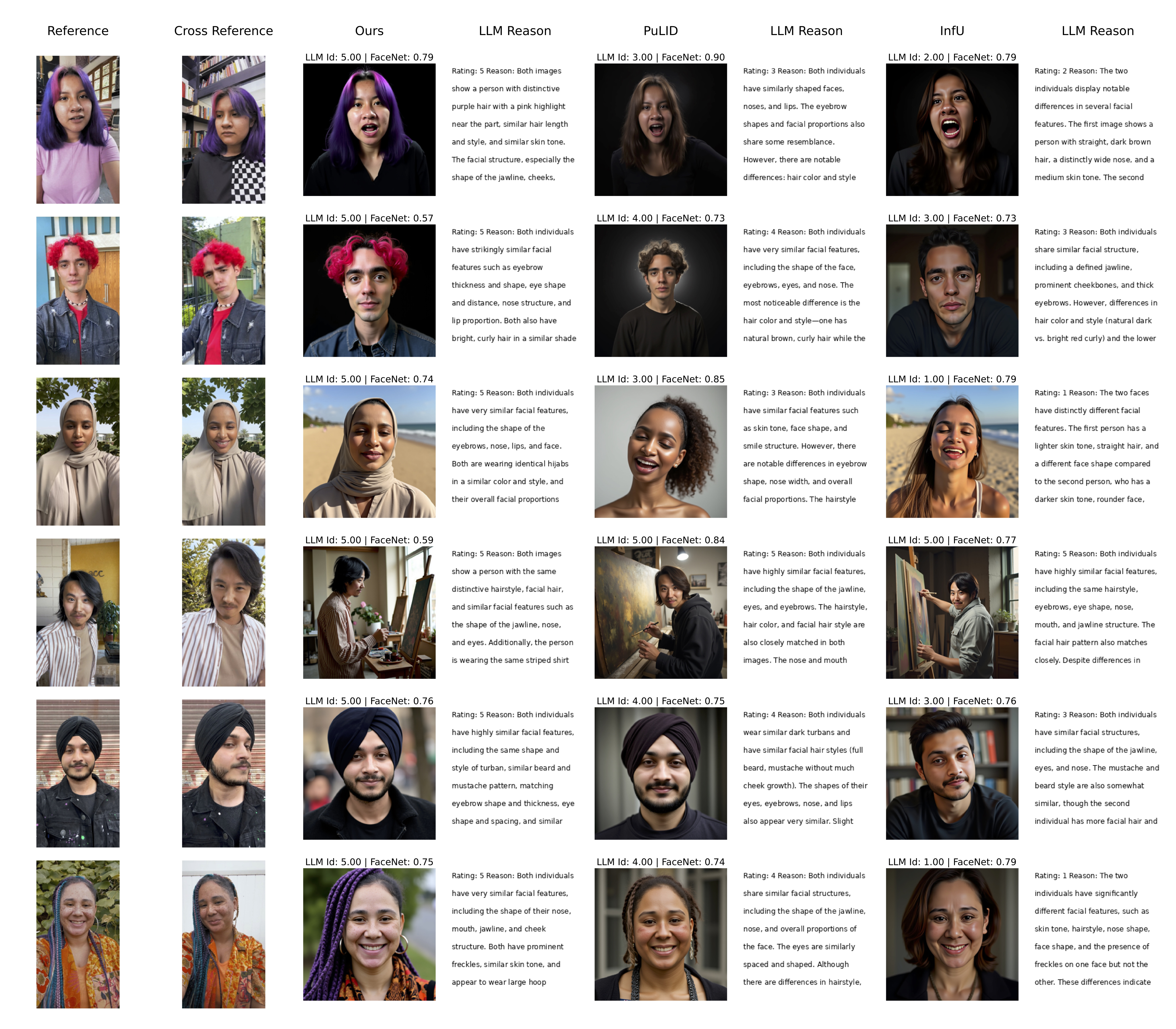}
    \caption{Qualitative comparison between LLM-based identity score and FaceNet Similarity.} The face crop from the reference image is used as input to the face adapter. The cross reference image is a shifted view of the same person and is used to compute identity scores.
    \label{fig:qual-llm-vs-facenet}
\end{figure}

We observe that face recognition models such as FaceNet~\citep{schroff2015facenet} and ArcFace~\citep{ArcFace}, commonly used for face similarity metrics, fail to account for variations in hairstyle, hair color, headwear, and head shape. We hypothesize that this limitation stems from their training setup, where the input face is tightly cropped to the size of $0.8\times$ of FFHQ~\citep{gal2021stylegan}), resulting in a narrow understanding of facial identity. Consequently, these models provide unreliable assessments of identity similarity in more holistic capture including hairstyle. To address this, we leverage large language models (LLMs) with strong multimodal capabilities—specifically ChatGPT-4o~\citep{ChatGPT4o}—to perform identity assessments. As illustrated in \cref{fig:qual-llm-vs-facenet}, the LLM accurately captures identity similarity in cases where FaceNet or ArcFace fails.

\section{LLM-Based Identity Evaluation Prompt}
\label{sec:llm-identity-prompt}

We use an LLM-as-a-judge to evaluate identity preservation in generated images. The LLM is given two face images and asked to judge whether they show the same person. The prompt is shown below:

\begin{quote}
\textbf{You are an expert in facial recognition. Given two images of faces, your task is to judge whether they show the same person.}

\textbf{Steps:}
\begin{enumerate}
    \item Carefully compare facial features like shape, eyes, nose, mouth, jawline, eyebrows, and overall proportions.
    \item Ignore lighting, angle, pose, worn accessories, or expressions.
    \item Rate identity similarity on a scale from 1 to 5:
    \begin{itemize}
        \item 5 = Definitely the same person
        \item 4 = Very likely the same person
        \item 3 = Possibly the same person
        \item 2 = Unlikely the same person
        \item 1 = Definitely not the same person
    \end{itemize}
    \item Explain your reasoning by referencing specific facial features.
\end{enumerate}

\textbf{Respond in the following format:}

Rating: \textless1--5\textgreater

Reason: \textless explanation\textgreater
\end{quote}

\section{LLM-Based Expression Evaluation Prompt}
\label{sec:llm-expression-prompt}

We use an LLM-as-a-judge to assess how well generated images reflect the intended facial expressions described in text prompts. The model is instructed to evaluate *only* the facial expression, disregarding pose, clothing, background, or lighting. The exact prompt is provided below:

\begin{quote}
\textbf{You are an expert judge tasked with evaluating how accurately a generated image captures the \textbf{facial expression} described in a text prompt. Your evaluation should focus \emph{only on the expression} (such as happiness, sadness, anger, surprise, etc.) and ignore other factors like the person's pose, clothing, background, or lighting.}

\textbf{Steps:}
\begin{enumerate}
    \item Carefully read the text prompt.
    \item Examine the facial expression in the generated image.
    \item Compare the image's expression to the description in the prompt.
    \item Rate the \textbf{expression fidelity} on a scale of 1 to 5:
    \begin{itemize}
        \item 5 = Perfect match.
        \item 4 = Mostly accurate with minor discrepancies.
        \item 3 = Partially matches but has clear inaccuracies.
        \item 2 = Mostly incorrect expression.
        \item 1 = Completely wrong expression.
    \end{itemize}
    \item Briefly explain \textbf{why} you gave this rating, pointing out specific facial features (mouth, eyes, brows, etc.) that contributed to your assessment.
\end{enumerate}

\textbf{Respond in the following format:}

Rating: \textless1--5\textgreater

Reason: \textless explanation \textgreater
\end{quote}

\section {Enabling expression, pose, and lighting control }
\begin{figure}[ht]
    \centering
    \includegraphics[width=\linewidth]{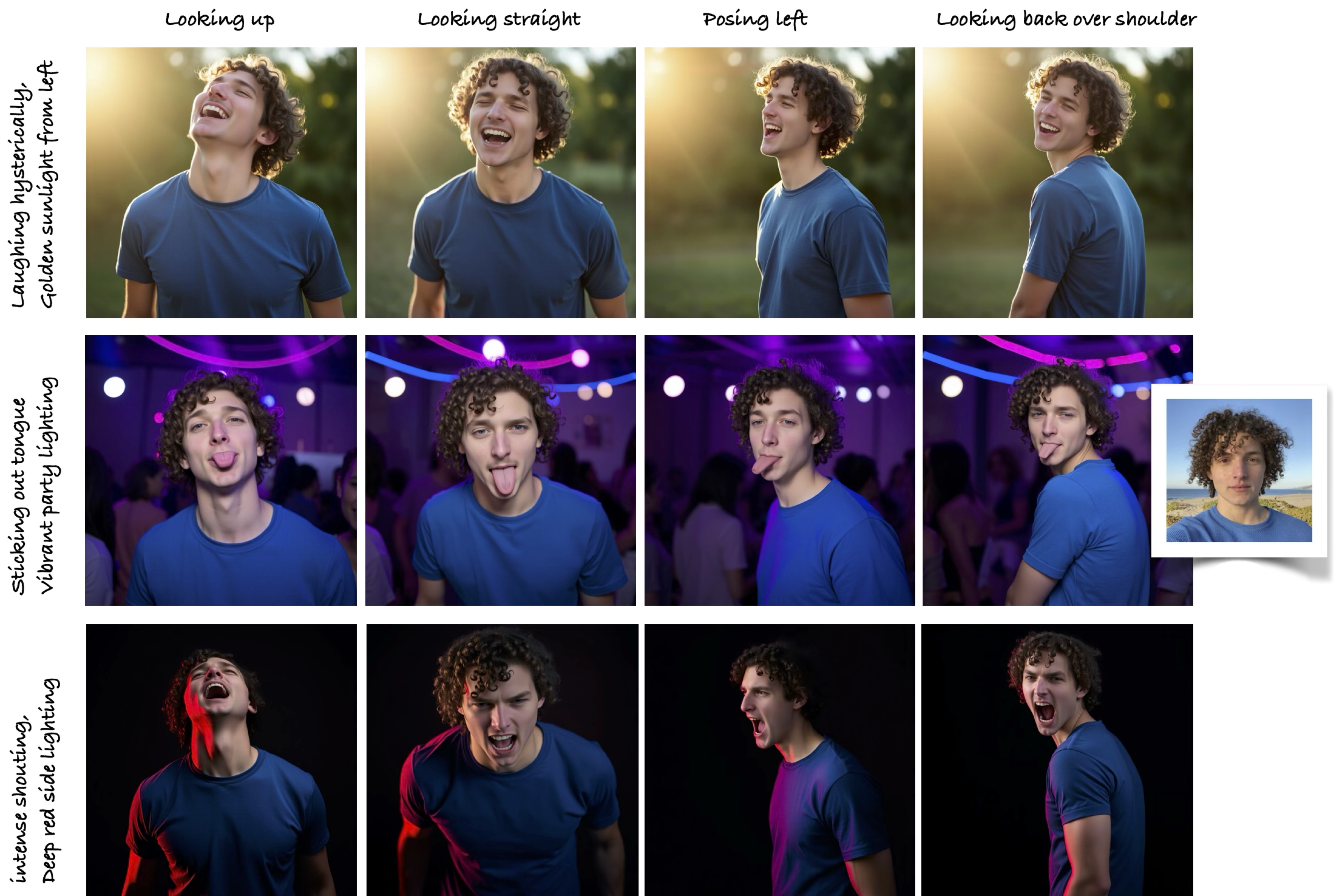}
    \caption{\textbf{Shortcut Rerouting enables to precisely control lighting, pose, and expression from text.} Adapter preservers from prior while utilizing only personalization related attributed from the reference image. Zoom in for best view.}
    \label{fig:supp:pose}
\end{figure}

In this section, we highlight the fine-grained control over pose, expression, and lighting enabled by our shortcut-rerouted training. In \ref{fig:supp:pose}, pose varies across columns while holding expression constant, and expression/lighting vary across rows with fixed pose. This disentangled control is achieved while faithfully preserving identity from the reference image.

\section{Additional Qualitative Results for Face Adapters}
% \begin{figure}[ht]
%     \centering
%     \includegraphics[width=\linewidth]{figures/src/x_app_figures_face_1015/merged_page_0.pdf}
%     \caption{\textbf{Qualitative comparison of different `face' adapters.} We test the adapters on their ability to edit and preserve wide variety facial expressions, pose, accessories, lighting, and activities. Zoom in for best view. }
%     \label{fig:apdx1}
% \end{figure}

\begin{figure}[ht]
    \centering
    \includegraphics[width=\linewidth]{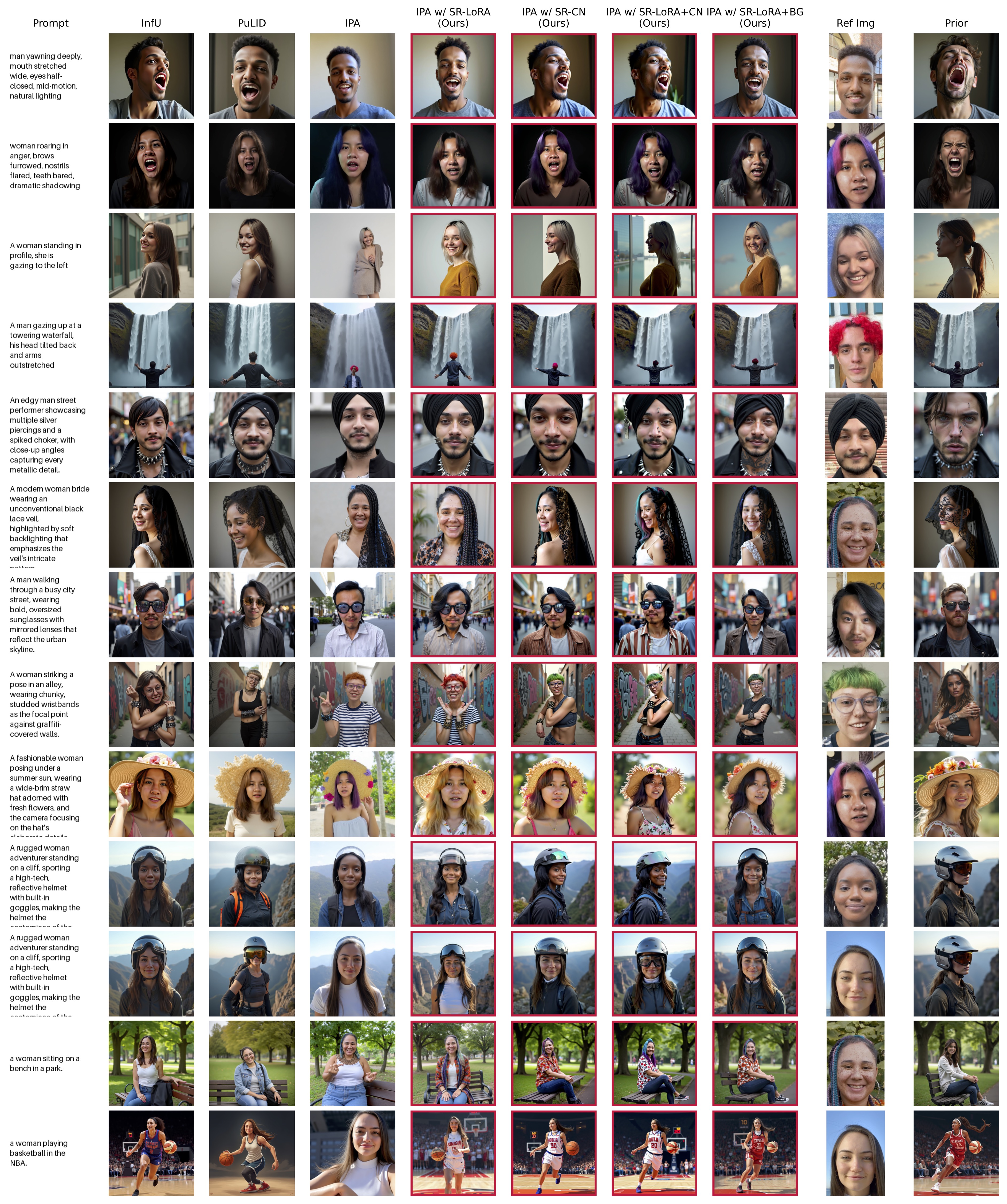}
    \caption{\textbf{Qualitative comparison of different ``face'' adapters.} We evaluate adapters on their ability to edit and preserve a wide variety of facial expressions, poses, accessories, lighting, and activities. \textbf{SR-LoRA} addresses quality degradation; \textbf{SR-CN} achieves better pose control; \textbf{SR-LoRA-BG} improves background and lighting consistency;  and using both SR-LoRA and SR-CN, namely \textbf{SR-LoRA-CN} effectively maintains pose while maintaining the quality; Zoom in for best view.}
    \label{fig:apdx2}
\end{figure}

\begin{figure}[ht]
    \centering
    \includegraphics[width=\linewidth]{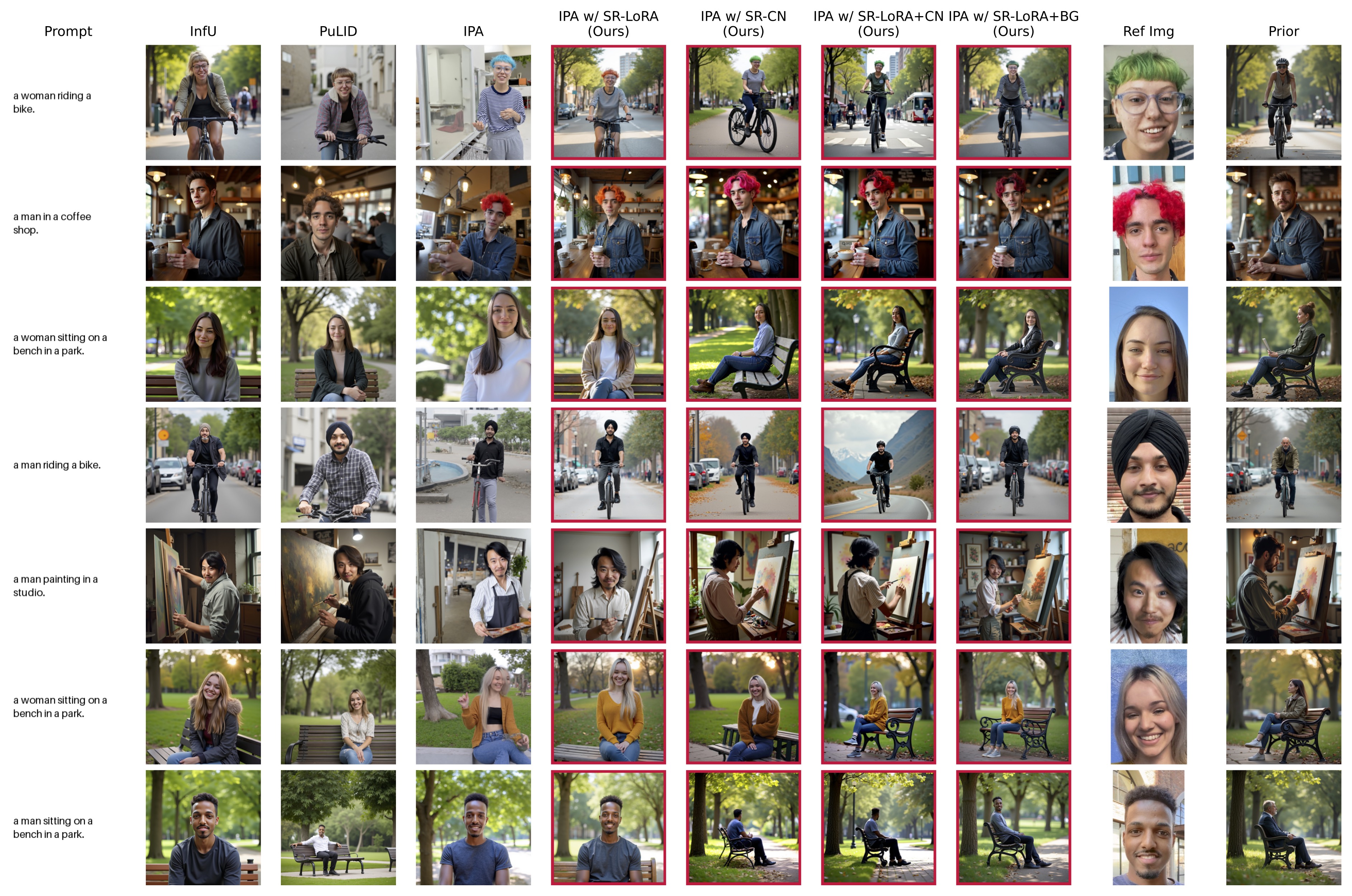}
    \caption{\textbf{Qualitative comparison of different ``face'' adapters.} We evaluate adapters on their ability to edit and preserve a wide variety of facial expressions, poses, accessories, lighting, and activities. \textbf{SR-LoRA} addresses quality degradation; \textbf{SR-CN} achieves better pose control; \textbf{SR-LoRA-BG} improves background and lighting consistency;  and using both SR-LoRA and SR-CN, namely \textbf{SR-LoRA-CN} effectively maintains pose while maintaining the quality; Zoom in for best view.}
\end{figure}

\section{Additional Qualitative Results for Fullbody Adapters}
% 
% \begin{figure}[ht]
%     \centering
%     \includegraphics[width=0.90\linewidth, trim=0 85 0 0, clip]{figures/x_app_fullbody_iter3/merged_selected_page_0.pdf}
%     \caption{\textbf{Qualitative comparison of different ``fullbody'' adapters.} Our approach consistently outperforms others by more accurately preserving body shape and enabling superior pose edit-ability.}
% \end{figure}

% Expression
% \begin{figure}[t]
%     \centering
%     \includegraphics[width=\linewidth]{figures/x_app_fullbody_iter3/merged_selected_page_1.jpg}
%     \caption{\textbf{Qualitative comparison of different ``fullbody'' adapters.} Our approach consistently outperforms others by more accurately preserving body shape and enabling superior pose edit-ability.}
% \end{figure}
% \clearpage

\section{Evaluation Prompts}

We randomly sample four identity images for each of the text prompts below, where {0} serves as a placeholder for the gender noun (e.g., man, woman).

\begin{table}[ht] \centering \resizebox{\textwidth}{!}{ \begin{tabular}{l}
\toprule
\textbf{Accessory} \\
\midrule
A distinguished \{0\} wearing an ornate, gold-trimmed monocle, paired with a classic black suit, where the monocle is illuminated under bright studio lights. \\
A fashionable \{0\} posing under a summer sun, wearing a wide-brim straw hat adorned with fresh flowers, and the camera focusing on the hat's elaborate details. \\
A glamorous \{0\} entertainer on stage, wearing extra-long, feathered earrings, with dramatic lighting accentuating their vivid colors and texture. \\
A glamorous \{0\} entertainer on stage, wearing extra-long, feathered earrings, with dramatic lighting accentuating their vivid colors and texture. \\
A modern \{0\} bride wearing an unconventional black lace veil, highlighted by soft backlighting that emphasizes the veil's intricate pattern. \\
A \{0\} striking a pose in an alley, wearing chunky, studded wristbands as the focal point against graffiti-covered walls. \\
A \{0\} walking through a busy city street, wearing bold, oversized sunglasses with mirrored lenses that reflect the urban skyline. \\
A retro-chic \{0\} fashion model in a vintage polka-dot scarf and matching gloves, with the patterned accessories taking center stage. \\
A rugged \{0\} adventurer standing on a cliff, sporting a high-tech, reflective helmet with built-in goggles, making the helmet the centerpiece of the shot. \\
A stylish \{0\} dancing in neon-lit surroundings, wearing knee-high lace-up boots sparkling with sequins, with the boots dominating the frame. \\
An edgy \{0\} street performer showcasing multiple silver piercings and a spiked choker, with close-up angles capturing every metallic detail. \\
\bottomrule
\end{tabular} } \end{table}

\begin{table}[ht] \resizebox{0.3\textwidth}{!}{ \begin{tabular}{l}
\toprule
\textbf{Activity} \\
\midrule
A \{0\} in a coffee shop. \\
A \{0\} in the office. \\
A \{0\} painting in a studio. \\
A \{0\} playing basketball in the NBA. \\
A \{0\} riding a bike. \\
A \{0\} sitting on a bench in a park. \\
\bottomrule
\end{tabular} } \end{table}

\begin{table}[ht] \centering \resizebox{\textwidth}{!}{ \begin{tabular}{l}
\toprule
\textbf{Expression} \\
\midrule
\{0\} biting lower lip nervously, slight frown, brows knit, awkward tension in the face \\
\{0\} blowing a kiss, lips puckered, eyebrows raised gently, soft lighting \\
\{0\} caught mid-expression between a laugh and a cry, watery eyes, twisted smile, bittersweet emotion \\
\{0\} crying intensely, eyebrows arched upward, mouth twisted in pain, eyes squeezed shut, raw emotion \\
\{0\} crying with one eye shut tighter than the other, mouth open mid-sob, flushed cheeks \\
\{0\} laughing hysterically with eyes shut tight and mouth wide open, cheeks raised, expressive lighting \\
\{0\} laughing with head tilted back, eyes closed, mouth wide open, pure joy on the face \\
\{0\} roaring in anger, brows furrowed, nostrils flared, teeth bared, dramatic shadowing \\
\{0\} screaming with eyes wide and jaw fully dropped, intense emotion on the face, dramatic lighting \\
\{0\} shocked with uneven brows, wide eyes, jaw slack, vivid emotion in facial pose \\
\{0\} shouting loudly, mouth wide, eyes intense, hands near face, motion blur \\
\{0\} smiling with eyes squeezed shut, mouth open in pure elation, cheeks lifted high \\
\{0\} smirking with head tilted slightly, one eyebrow raised, subtle attitude in the eyes \\
\{0\} snarling with clenched teeth, nose scrunched, eyes narrowed in aggression \\
\{0\} sneering in disgust, nose wrinkled, upper lip curled, one eye slightly squinted, gritty atmosphere \\
\{0\} sticking out tongue in a mocking expression, playful eyes, raised eyebrow, casual setting \\
\{0\} stunned with jaw dropped, eyebrows raised high, eyes wide open, sharp lighting \\
\{0\} wincing in pain with eyes tightly shut, lips twisted, brow tense, expressive emotion \\
\{0\} winking with a mischievous grin, one eye squinting and shut, playful mood \\
\{0\} winking with a mischievous grin, one eye squinting, playful mood \\
\{0\} yawning deeply, mouth stretched wide, eyes half-closed, mid-motion, natural lighting \\
\bottomrule
\end{tabular} } \end{table}

\begin{table}[ht] \resizebox{0.5\textwidth}{!}{ \begin{tabular}{l}
\toprule
\textbf{Lighting} \\
\midrule
A \{0\} dancing under shifting multicolor disco lights \\
A \{0\} gazing sideways, face partially lit by peach morning light \\
A \{0\} in a dynamic pose under cold cyan lighting \\
A \{0\} in crouched pose under deep indigo overhead spotlight \\
A \{0\} leaning against a wall, lit from below with greenish hue \\
A \{0\} leaning forward into cool violet side light \\
A \{0\} mid-motion, under red and purple colored strobe lights \\
A \{0\} reaching upward under golden sunrise rays \\
A \{0\} reclining on couch under dusty rose lighting \\
A \{0\} sitting cross-legged, lit with emerald green hue \\
A \{0\} sitting sideways under diffused teal lighting \\
A \{0\} standing tall, shadow cast long under low amber light \\
A \{0\} standing under soft orange sunset light, profile facing right \\
A \{0\} turning back toward camera, lit from behind with white light \\
A \{0\} under warm candlelight, hands resting on lap \\
A \{0\} walking into a warm golden spotlight from the side \\
A \{0\} with arms crossed, under vibrant magenta rim lighting \\
A \{0\} with hands on hips, under bright orange studio lights \\
A \{0\} with head tilted back, lit by soft lavender haze \\
A \{0\} with one hand in pocket, standing in blue-tinted window light \\
\bottomrule
\end{tabular} } \end{table}

% \subsection*{Pose}

\begin{table}[ht] \centering \resizebox{\textwidth}{!}{ \begin{tabular}{l}
\toprule
\textbf{Pose} \\
\midrule
A \{0\} gazing up at a towering waterfall, their head tilted back and arms outstretched \\
A \{0\} kneeling on one knee in a dramatic, torchlit cave, leaning forward with eyes fixed on the horizon, exuding focus and intensity. \\
A \{0\} leaning against a graffiti-covered wall, hands in pockets, shoulders slightly slouched, glancing away with a laid-back expression. \\
A \{0\} seated cross-legged on a wooden floor, hands resting on knees, gazing slightly downward in meditative calm. \\
A \{0\} standing in profile, they are gazing to the left \\
A \{0\} standing in profile, they are gazing to the left \\
A \{0\} standing tall, their hands on their hips, and their chin held high, in front of a modern office backdrop. \\
A \{0\} standing tall, with their hands on their hips, and their chin held high, in front of a modern office backdrop. \\
A stylish \{0\} turning to glance over their right shoulder, with their profile in partial silhouette against a glowing city skyline at sunset. \\
A confident \{0\} facing forward, looking directly into the camera lens with a poised, straight-backed posture, set against a clean studio background. \\
\bottomrule
\end{tabular} } \end{table}

\end{document}